\documentclass[10pt,twocolumn,letterpaper]{article}

\usepackage[pagenumbers]{cvpr} 

\usepackage[dvipsnames]{xcolor}

\newcommand{\inlinesection}[1]{\vspace{0.05cm} \noindent {\bf #1}}
\newcommand{\titlecaption}[2]{\caption{\small \textbf{#1.}\xspace#2}}

\definecolor{cvprblue}{rgb}{0.21,0.49,0.74}
\usepackage[pagebackref,breaklinks,colorlinks,citecolor=cvprblue]{hyperref}

\usepackage{amsmath}
\usepackage{amssymb}
\usepackage{bm,mathtools}

\def\eqref#1{equation~\ref{#1}}

\def\1{\bm{1}}

\def\rvz{{\mathbf{z}}}

\def\rmM{{\mathbf{M}}}

\DeclareMathAlphabet{\mathsfit}{\encodingdefault}{\sfdefault}{m}{sl}
\SetMathAlphabet{\mathsfit}{bold}{\encodingdefault}{\sfdefault}{bx}{n}

\usepackage{url}
\usepackage{graphicx}
\usepackage{adjustbox,booktabs,multirow}
\usepackage{caption}
\usepackage{subcaption}
\usepackage{pifont}
\usepackage{xspace}
\usepackage{tabularx}
\usepackage{wrapfig}
\usepackage{xcolor}

\usepackage{tikz}
\usetikzlibrary{shapes}
\usetikzlibrary{calc}
\usetikzlibrary{backgrounds}
\usetikzlibrary{positioning}
\usetikzlibrary{arrows}
\usetikzlibrary{arrows.meta}
\usetikzlibrary{decorations.text}    
\usetikzlibrary{decorations.pathreplacing}
\usetikzlibrary{fit}
\usetikzlibrary{spy}
\usetikzlibrary{patterns}

\usepackage{pgfplotstable,filecontents}
\usepgfplotslibrary{colorbrewer}
\pgfplotsset{compat = 1.15}

\usepackage{paralist}

\definecolor{codegreen}{rgb}{0,0.6,0}
\definecolor{codegray}{rgb}{0.5,0.5,0.5}
\definecolor{codepurple}{rgb}{0.58,0,0.82}
\definecolor{backcolour}{rgb}{0.95,0.95,0.92}
\definecolor{codered}{rgb}{0.89,0.4,.45}

\usepackage{listings}
\lstdefinestyle{mystyle}{
    backgroundcolor=\color{backcolour},   
    commentstyle=\color{codegreen},
    keywordstyle=\color{codered},
    numberstyle=\tiny\color{codegray},
    stringstyle=\color{codepurple},
    basicstyle=\ttfamily\footnotesize,
    breakatwhitespace=false,         
    breaklines=true,                 
    captionpos=b,                    
    keepspaces=true,                 
    numbers=left,                    
    numbersep=5pt,                  
    showspaces=false,                
    showstringspaces=false,
    showtabs=false,                  
    tabsize=2
}

\definecolor{turquoise}{cmyk}{0.65,0,0.1,0.3}
\definecolor{purple}{rgb}{0.65,0,0.65}
\definecolor{dark_green}{rgb}{0, 0.5, 0}
\definecolor{orange}{rgb}{0.8, 0.6, 0.2}
\definecolor{red}{rgb}{0.8, 0.2, 0.2}
\definecolor{darkred}{rgb}{0.6, 0.1, 0.05}
\definecolor{blueish}{rgb}{0.0, 0.3, .6}
\definecolor{light_gray}{rgb}{0.7, 0.7, .7}
\definecolor{pink}{rgb}{1, 0, 1}
\definecolor{greyblue}{rgb}{0.25, 0.25, 1}

\newcommand{\oursu}{\text{SV3D}^u}
\newcommand{\oursc}{\text{SV3D}^c}
\newcommand{\oursp}{\text{SV3D}^p}

\usepackage{graphicx,multirow}

\title{SV3D: Novel Multi-view Synthesis and 3D Generation from a Single Image using Latent Video Diffusion}

\author{Vikram Voleti$^*$ \quad Chun-Han Yao$^*$ \quad Mark Boss$^*$ \quad Adam Letts \quad David Pankratz \\ \quad Dmitry Tochilkin \quad Christian Laforte \quad Robin Rombach \quad Varun Jampani$^*$ \\ \\ Stability AI}

\providecommand{\impath}[1]{}

\newcommand{\illumination}{
\begin{figure}[htb!]

    \begin{tabular}{@{}cc@{}}
        \includegraphics[width=0.45\linewidth, trim={20 100 10 120},clip]{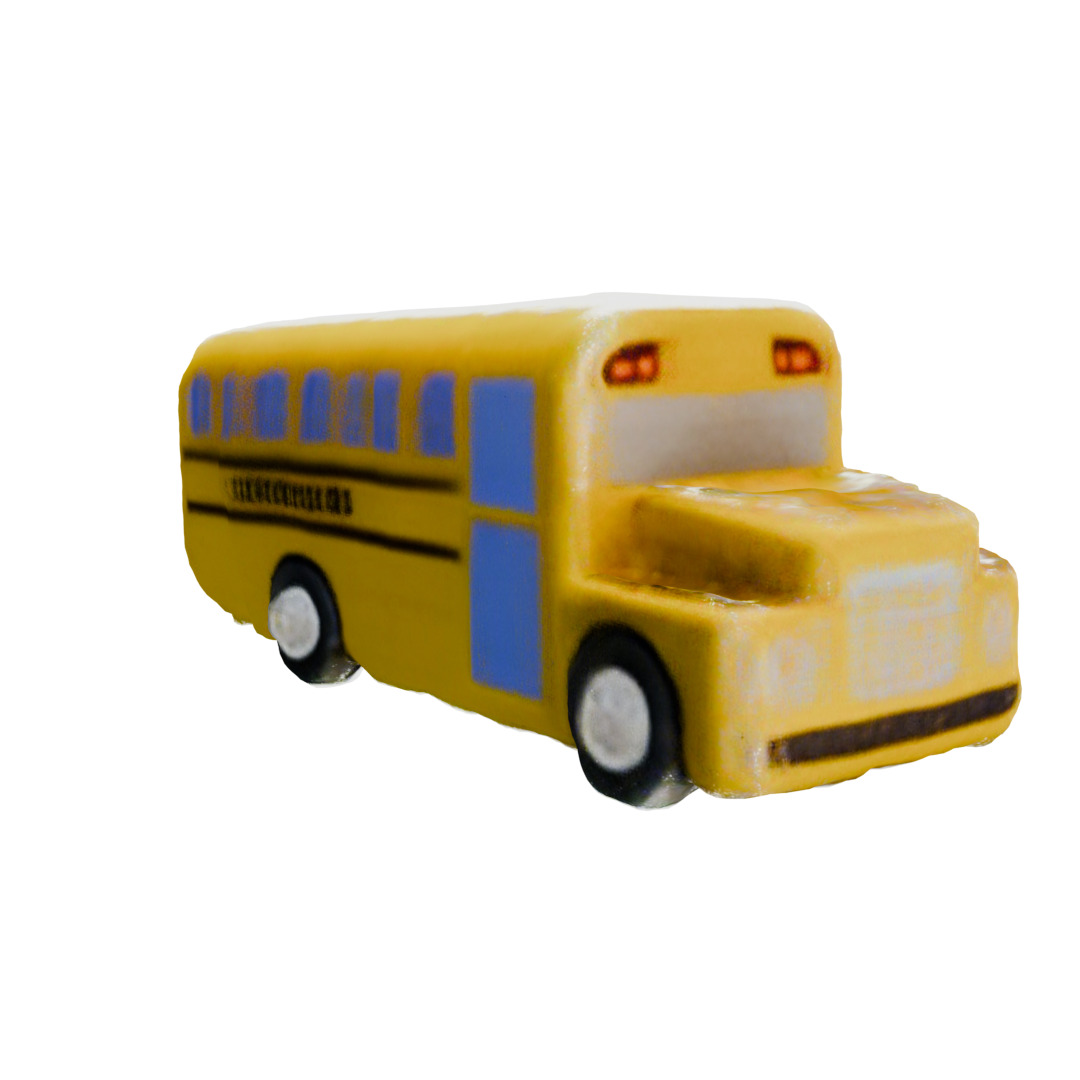} &
        \includegraphics[width=0.45\linewidth, trim={20 100 10 120},clip]{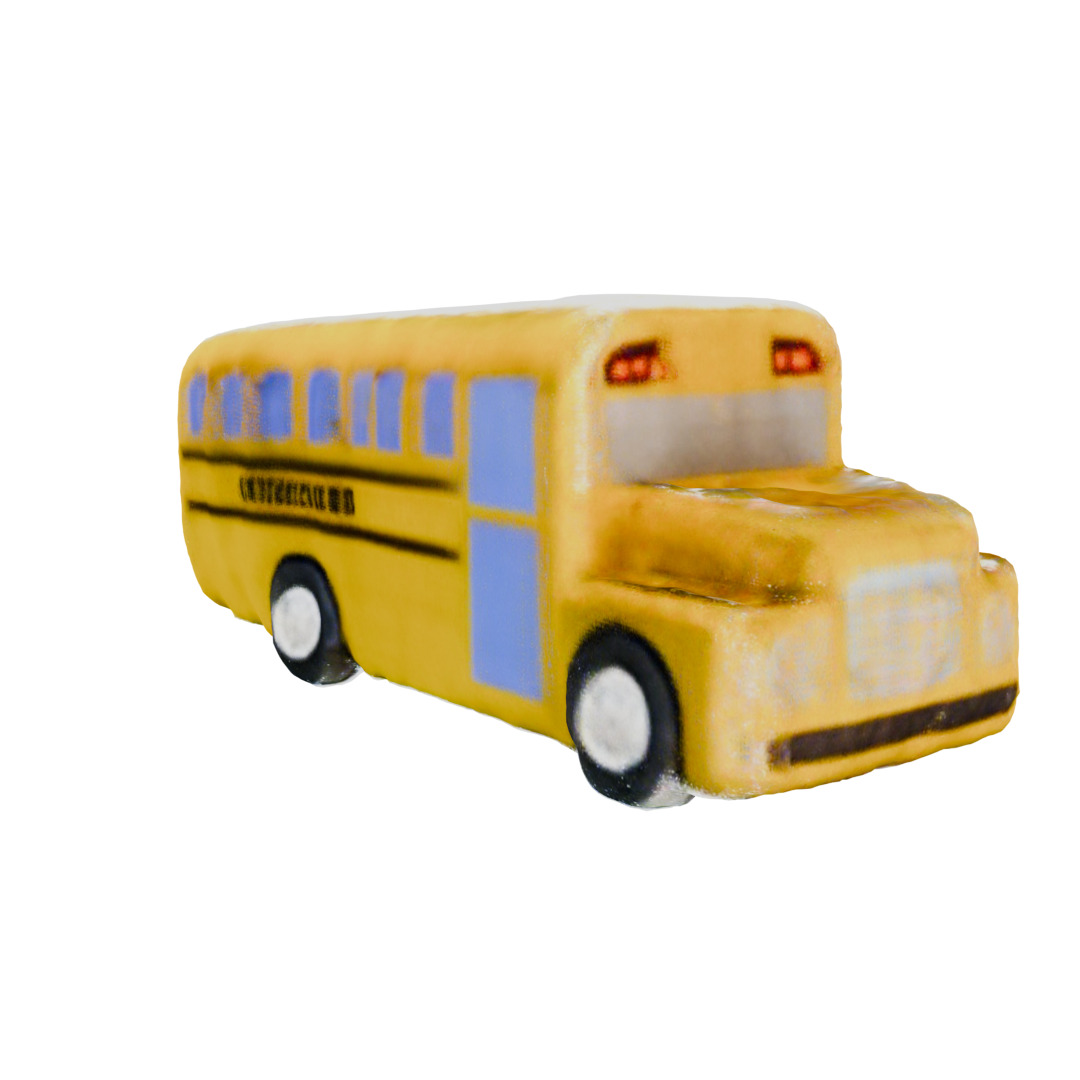} \\
        \tiny Constant Illumination & \tiny SGs Illumination \\
    \end{tabular}
    \titlecaption{Constant vs. SGs Illumination}{
        Notice that our SGs-based reconstructions do not exhibit darkening on the side of the bus, which enables easier and more convincing relighting for downstream applications.
    }
    \label{fig:diffuse_vs_sgs}
\end{figure}
}

\newcommand{\orbitsphotosds}{
\begin{figure}[htb!]
\centering
\includegraphics[width=.8\linewidth]{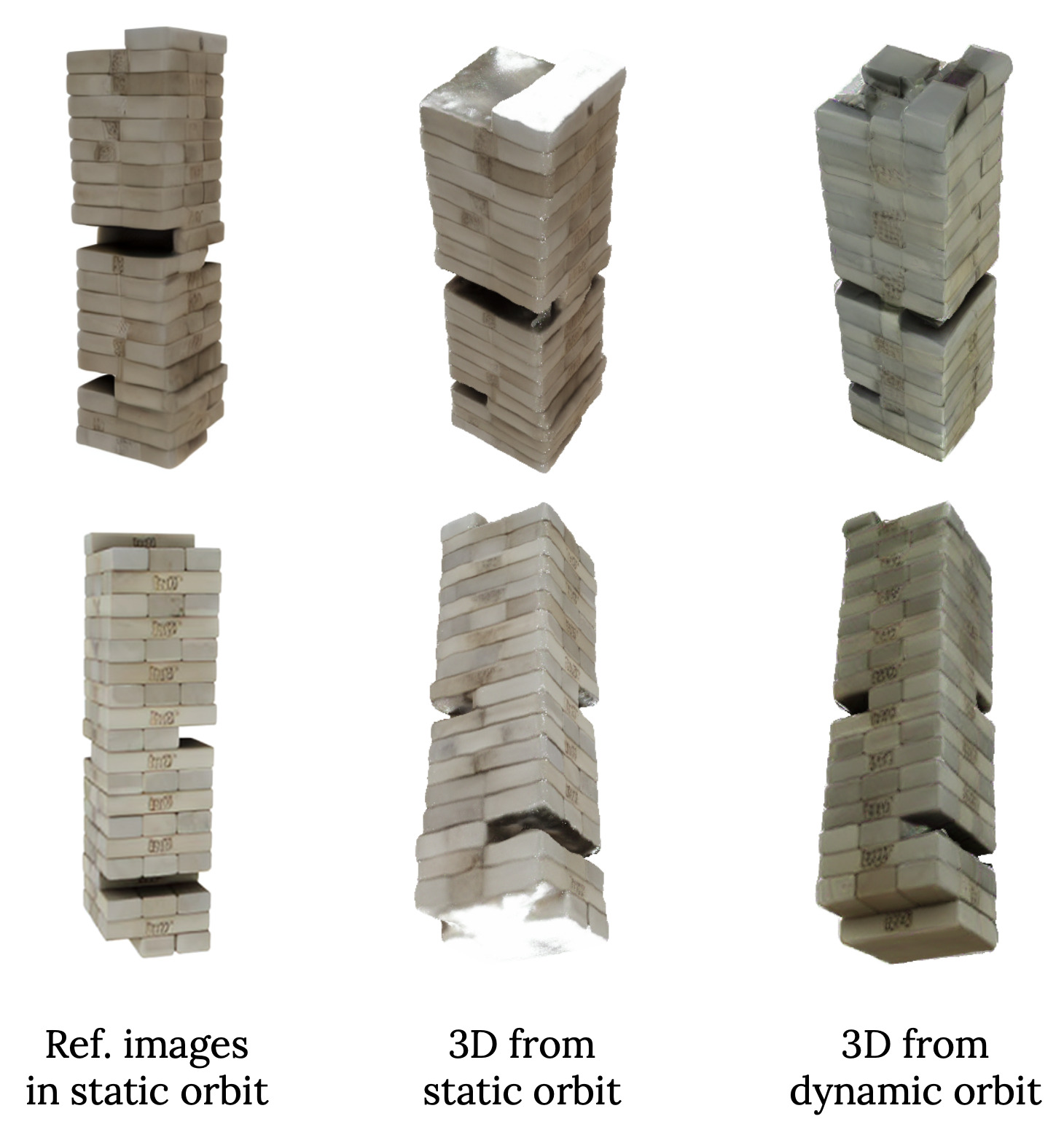}
\titlecaption{Influence of Training Orbits}{
    We show that using a dynamic orbit is crucial to 3D generations that are complete from diverse views.
}
\label{fig:orbits}
\end{figure}
\begin{figure}[htb!]
    \centering
    \includegraphics[width=0.9\linewidth, trim={0 0 0 20},clip]{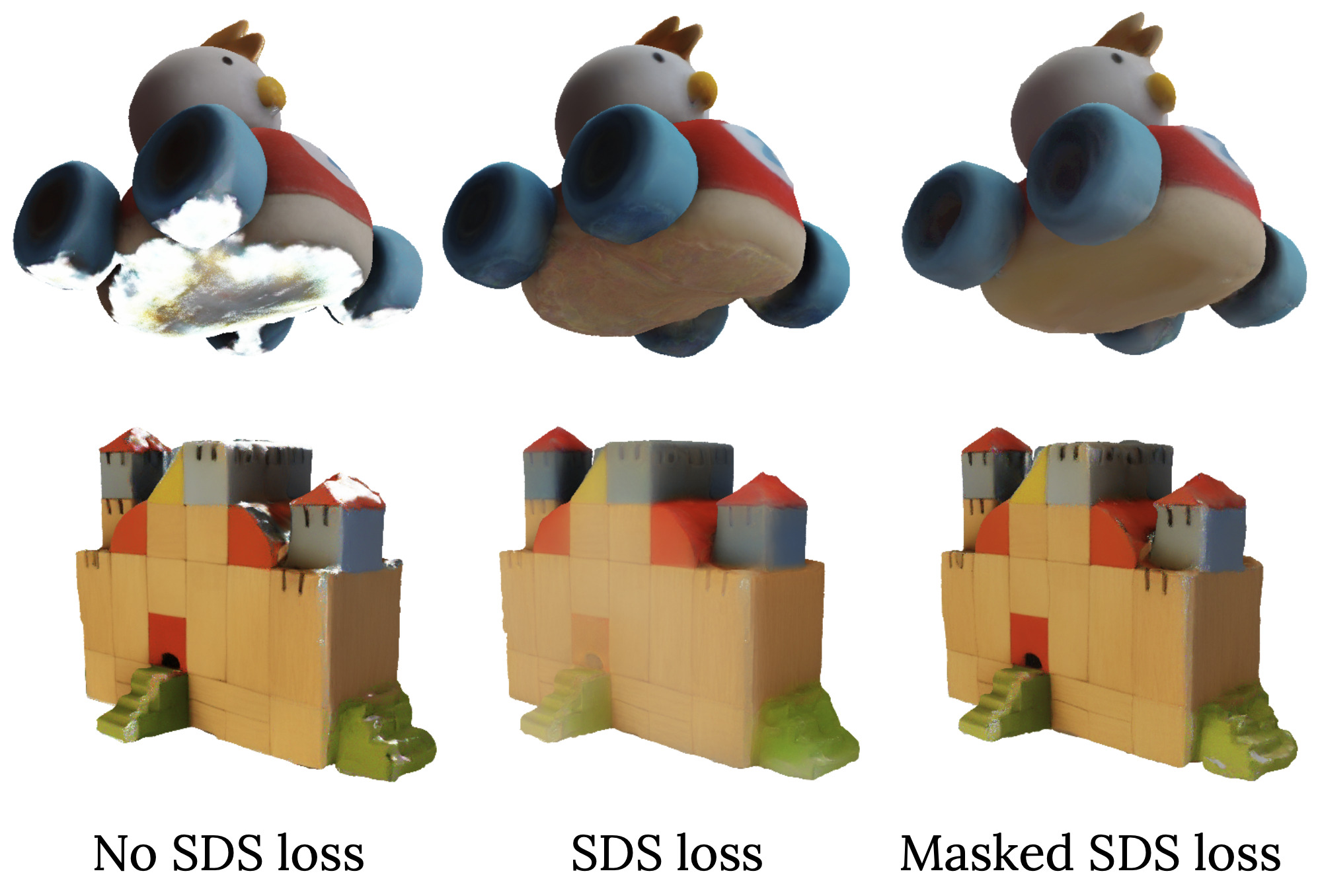}
    \titlecaption{Influence of SDS}{
        Using our masked SDS loss, we are able to fill in unseen surfaces in the training orbit, producing a cleaner result without oversaturation or blurry artifacts caused by naive SDS.
    }
    \label{fig:photo_sds}
\end{figure}
}

\newcommand{\baselinethreed}{
\begin{figure*}[t]
    \centering
    \includegraphics[width=0.9\linewidth]{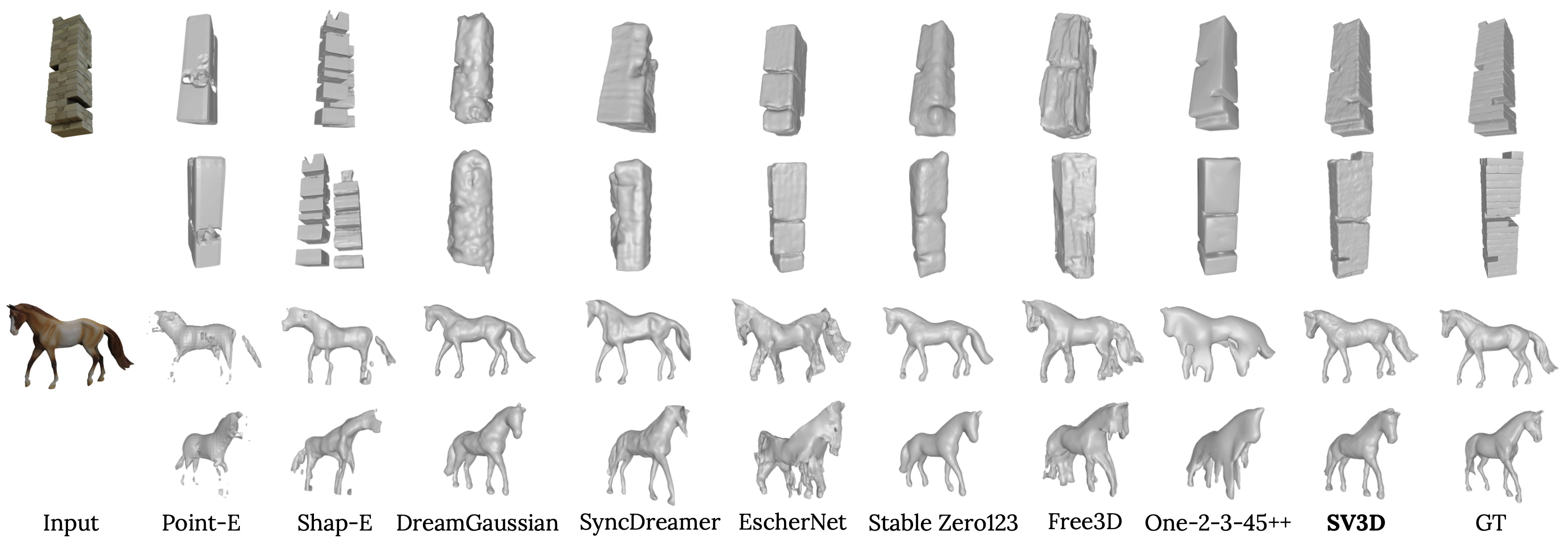}
    \titlecaption{Visual Comparison of 3D Meshes}{
        For each object, we show the conditioning image (left), and the output meshes rendered in two different views. Our generated meshes are more detailed, faithful to input images, and consistent in 3D. This demonstrates the quality of novel multi-view synthesis by our SV3D model.
    }
\label{fig:3d_baselines}
\end{figure*}
}

\newcommand{\realthreed}{
\begin{figure*}[t]
    \centering
    \includegraphics[width=0.9\linewidth]{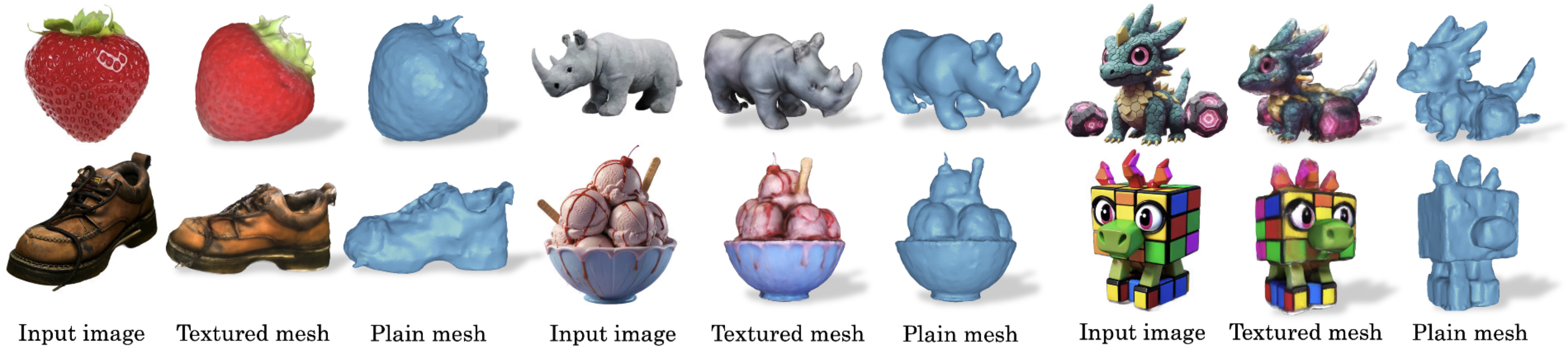}
    \titlecaption{Real-World 3D Results}{
        Notice the accurate shape and details in our reconstructions even from the diverse images in-the-wild.
    }
\label{fig:3d_real}
\end{figure*}
}

\newcommand{\omniobjectthreedsupp}{
\begin{figure*}[!]
    \centering
    \includegraphics[width=\linewidth]{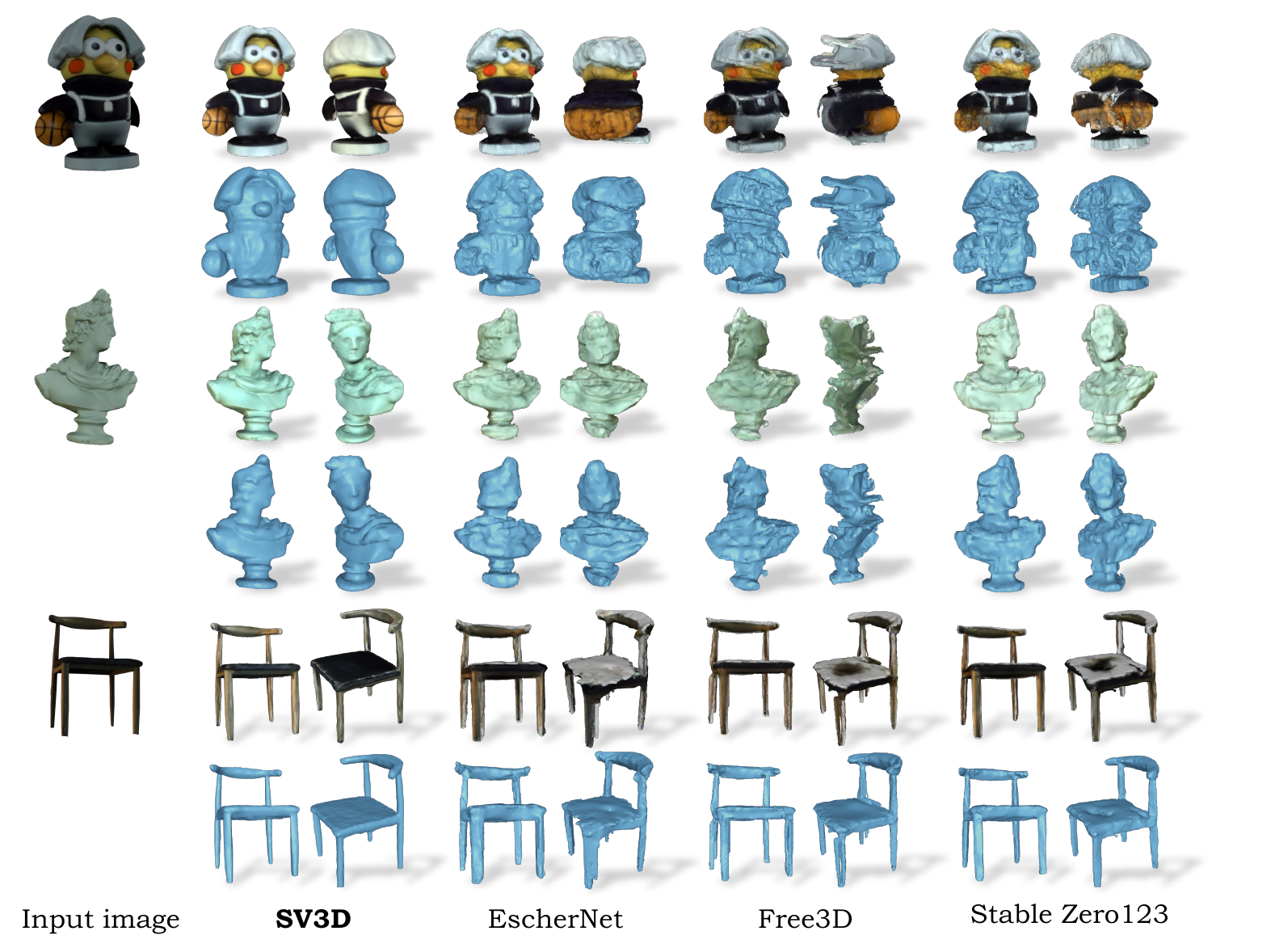}
    \titlecaption{Mesh Results on OmniObject3D~\cite{wu2023omniobject3d}}{
        Notice the accurate shape and details in our reconstructions even from the diverse images.
    }
\label{fig:3d_omni_supp}
\end{figure*}
}

\newcommand{\realthreedsupp}{
\begin{figure*}[!]
    \centering
    \includegraphics[width=\linewidth]{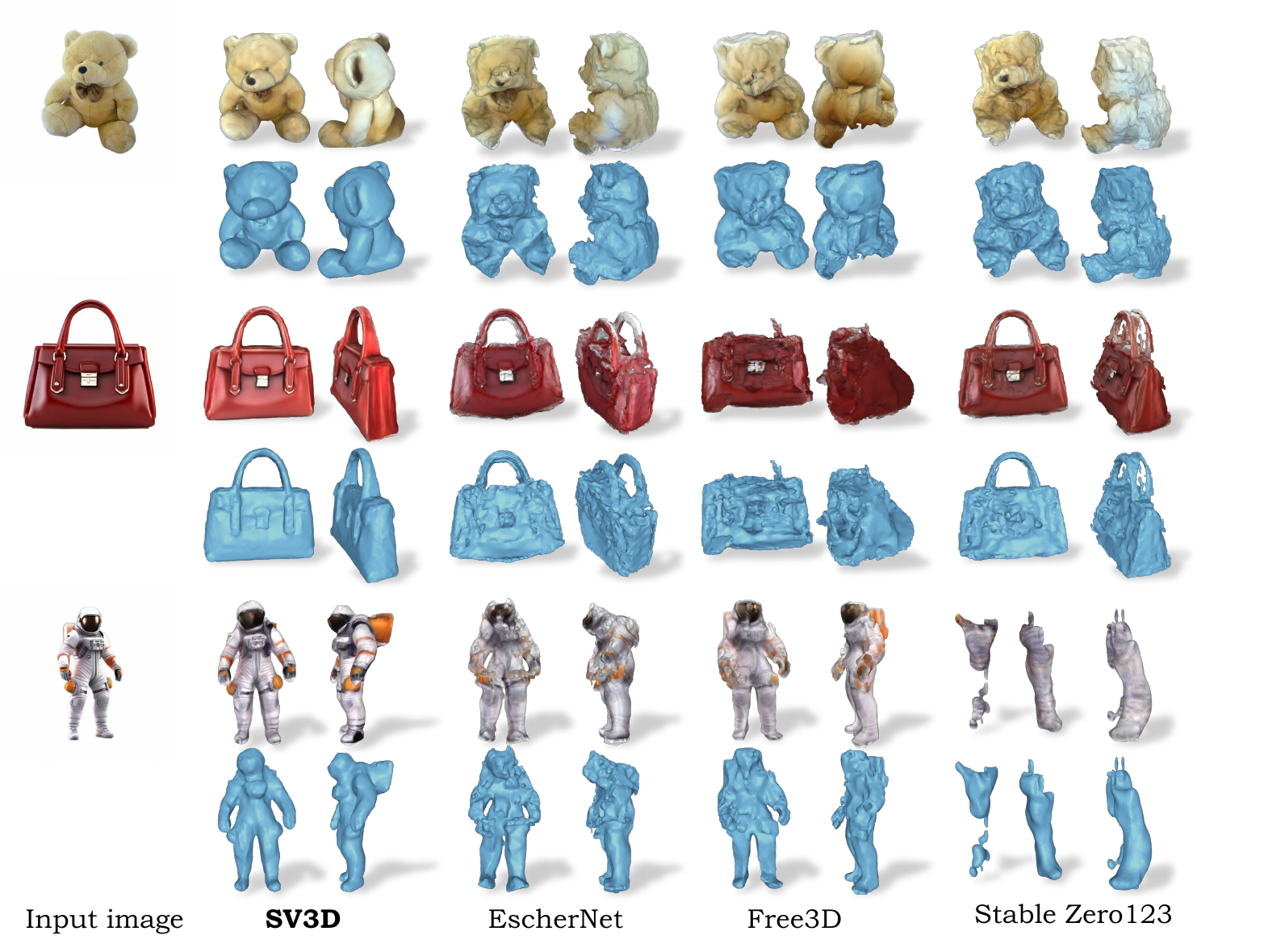}
    \titlecaption{Mesh Results on Real-World images}{
        Notice the accurate shape and details in our reconstructions even from the diverse images in-the-wild.
    }
\label{fig:3d_real_supp}
\end{figure*}
}

\newcommand{\omniobjectnvssupp}{
\begin{figure*}[!]
    \centering
    \includegraphics[width=\linewidth]{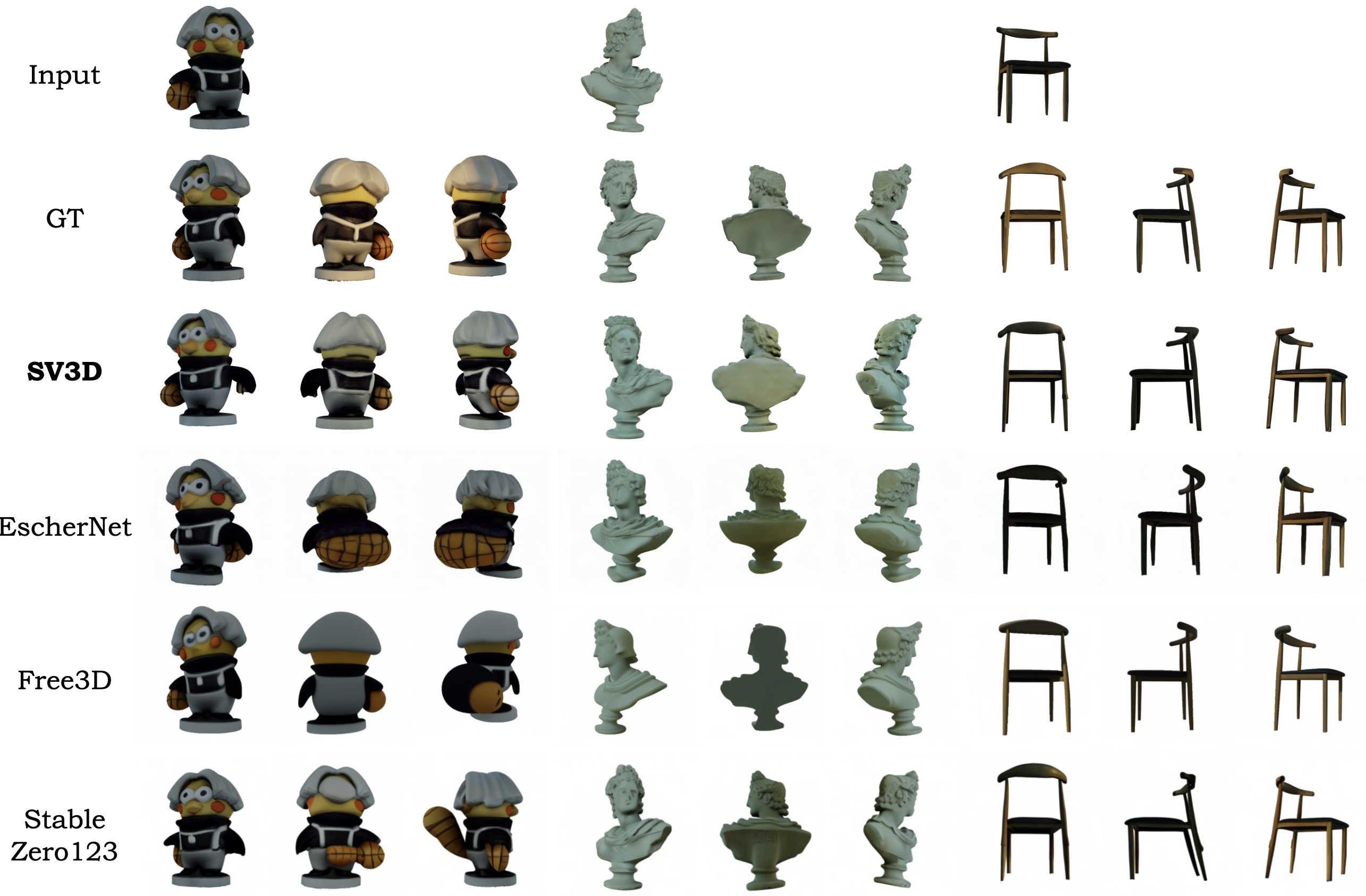}
    \titlecaption{NVS Results on OmniObject3D~\cite{wu2023omniobject3d}}{
        Notice the consistent color, geometry, and pose in SV3D NVS outputs compared to prior works.
    }
\label{fig:nvs_omni_supp}
\end{figure*}
}

\newcommand{\realnvssupp}{
\begin{figure*}[!]
    \centering
    \includegraphics[width=\linewidth]{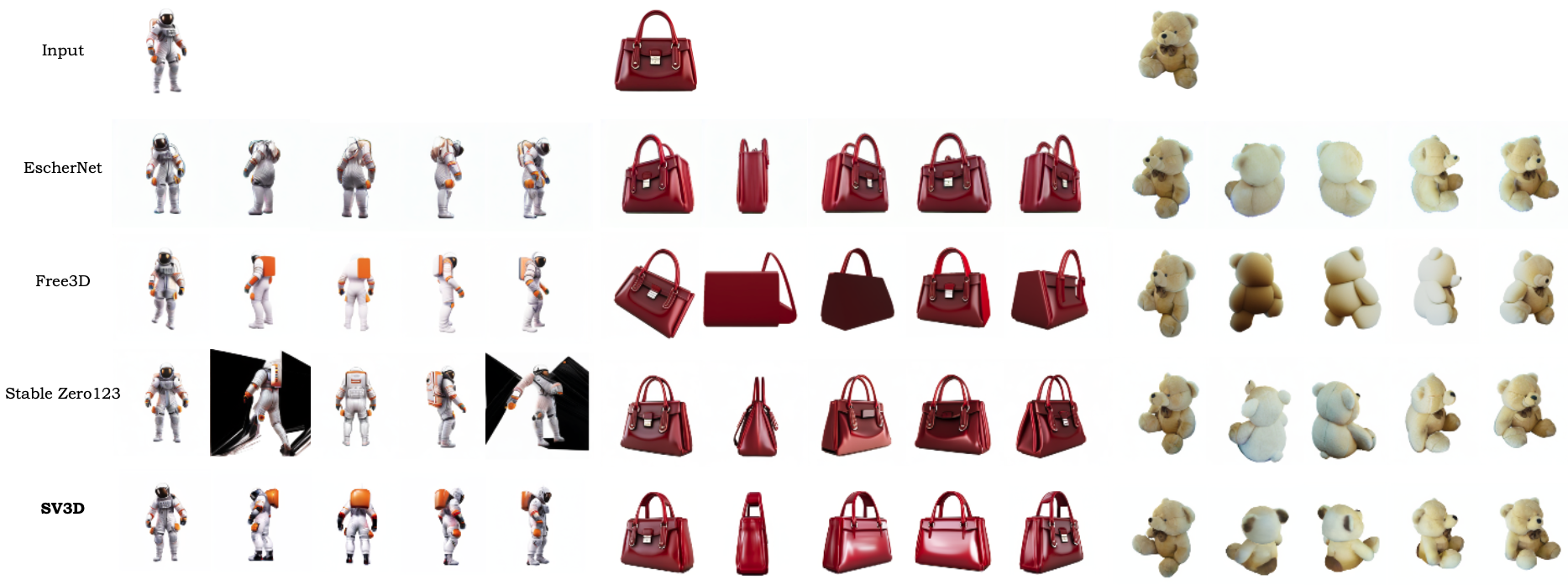}
    \titlecaption{NVS Results on Real-World images}{
        Notice the consistent color, geometry, and pose in SV3D NVS outputs compared to prior works.
    }
\label{fig:nvs_real_supp}
\end{figure*}
}

\newcommand{\NVScomp}{
\begin{figure*}[t]
    \centering
    \includegraphics[width=\linewidth]{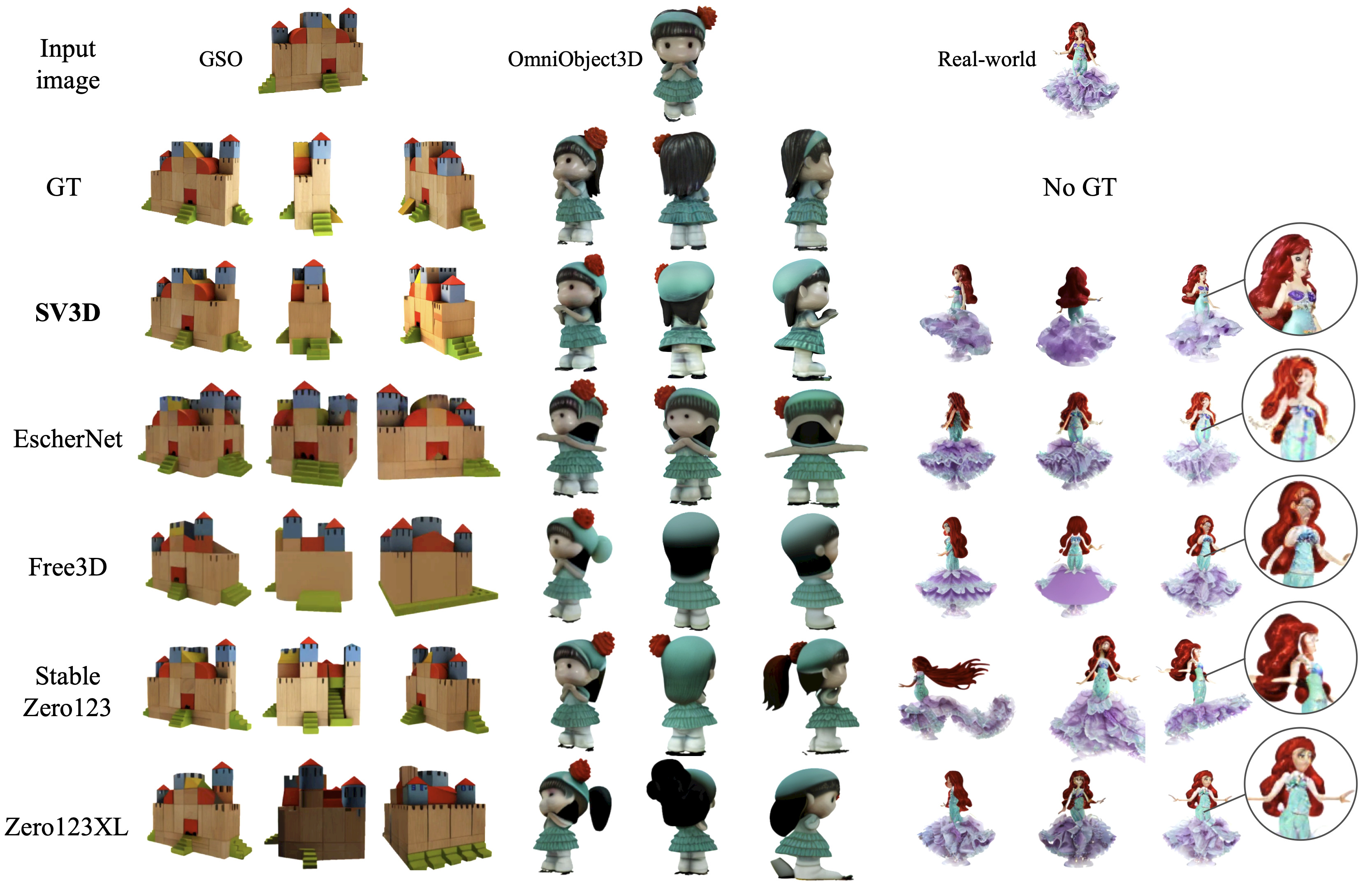}
    \titlecaption{Visual Comparison of Novel Multi-view Synthesis}{
        SV3D is able to generate novel multi-views that are more detailed, faithful to the conditioning image, and multi-view consistent compared to the prior works.
    }
    \label{fig:nvs_comp}
\end{figure*}
}

\begin{document}

\twocolumn[{%
    \renewcommand\twocolumn[1][]{#1}%
    \maketitle
    
    
    \begin{center}
    \newcommand{\whitebox}{\hfill\textcolor{white}{\rule[1mm]{1.8mm}{2.8mm}}\hfill}
\newcommand{\filmbox}[1]{%
    \setlength{\fboxsep}{0pt}%
    \colorbox{black}{%
        \begin{minipage}{3.2cm}
            \rule{0mm}{4.8mm}\whitebox\whitebox\whitebox\whitebox\whitebox%
            \whitebox\whitebox\whitebox\whitebox\null\\%
            \null\hfill\includegraphics[width=3cm]{#1}\hfill\null\\[1mm]%
            \null\whitebox\whitebox\whitebox\whitebox\whitebox%
            \whitebox\whitebox\whitebox\whitebox\null
        \end{minipage}}}
\begin{tikzpicture}[
    encoder/.style={trapezium, trapezium angle=60, text width=0.3cm, text height=0.4cm, shape border rotate=0, draw=#1, fill=#1!20, line width=1.5pt, inner sep=0.1cm, rounded corners=1pt},
    decoder/.style={trapezium, trapezium angle=60, text width=0.3cm, text height=0.4cm, shape border rotate=180, draw=#1, fill=#1!20, line width=1.5pt, inner sep=0.1cm, rounded corners=1pt},
]
    
    \node (film_strip) {\resizebox{0.3\textwidth}{!}{\filmbox{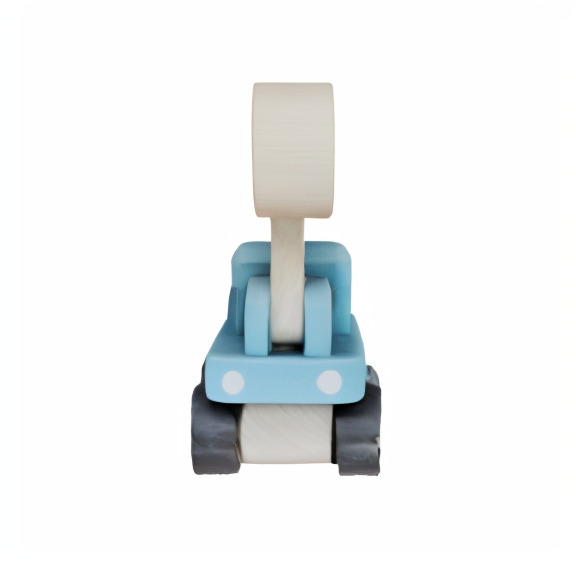}\filmbox{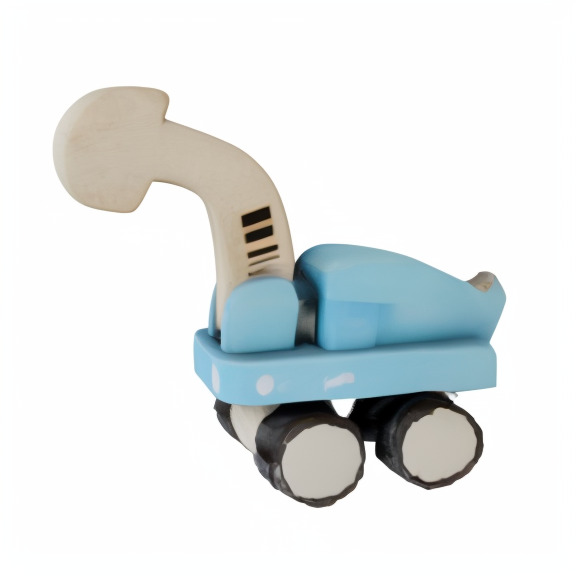}\filmbox{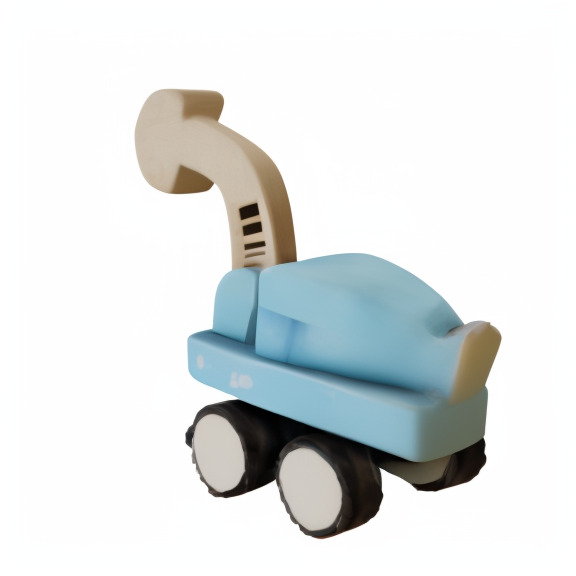}\filmbox{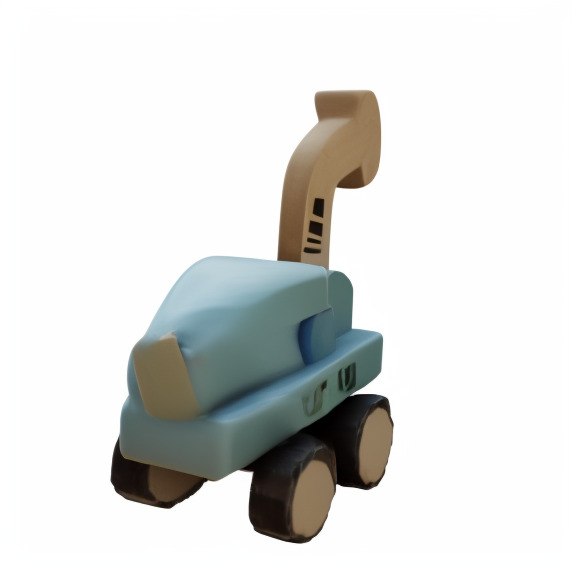}\filmbox{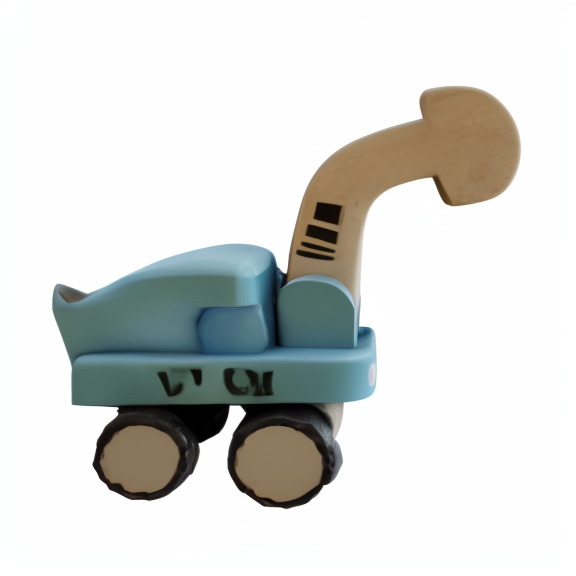}}};

    \node[above right=0.1cm of film_strip.north west, anchor=south west, draw=black, line width=1pt] (input_img) {\includegraphics[width=0.1\textwidth]{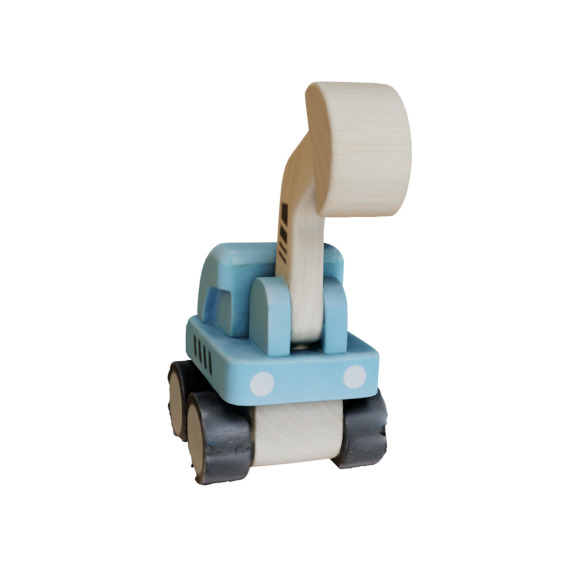}};
    \node[right=0.6cm of input_img, anchor=south, rotate=-90, encoder={teal}] (encoder) {};
    \node[right=0cm of encoder.north, anchor=south, decoder={teal}, rotate=-90] (decoder) {};
    \node[below=0.4cm of encoder.north] (sv3d_label) {\tiny SV3D};

    \node[right=1.2cm of $(film_strip.east)!0.5!(decoder.south)$, yshift=0.15cm] (nerf_viz) {\includegraphics[width=0.2\textwidth]{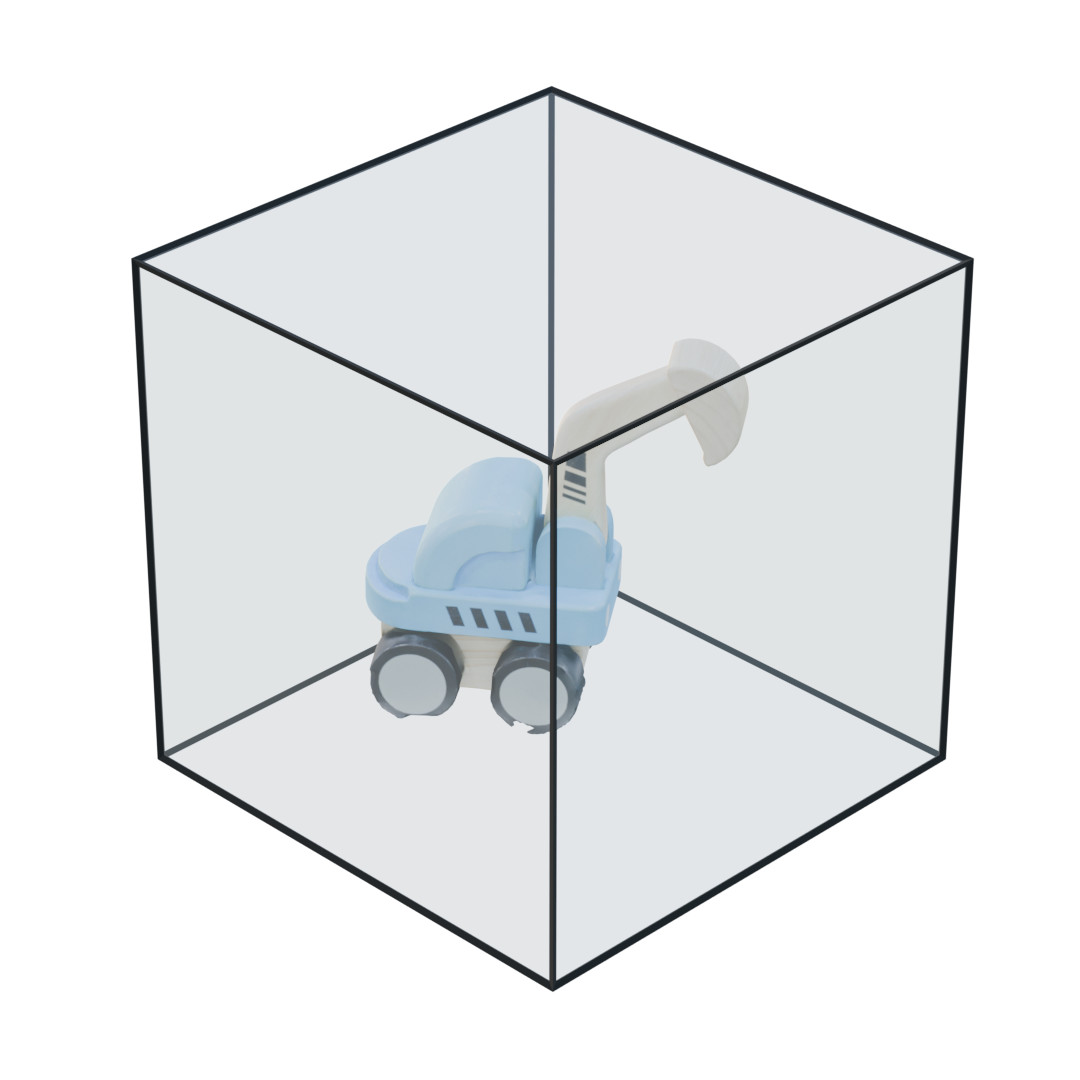}};
    \node[right=1.2cm of $(film_strip.east)!0.5!(decoder.south)$, below=0.2cm of nerf_viz] (opt_label) {\scriptsize 3D Optimization};
    \node[right=-0.2cm of nerf_viz, yshift=-0.1cm] (examples) {\includegraphics[width=0.35\textwidth]{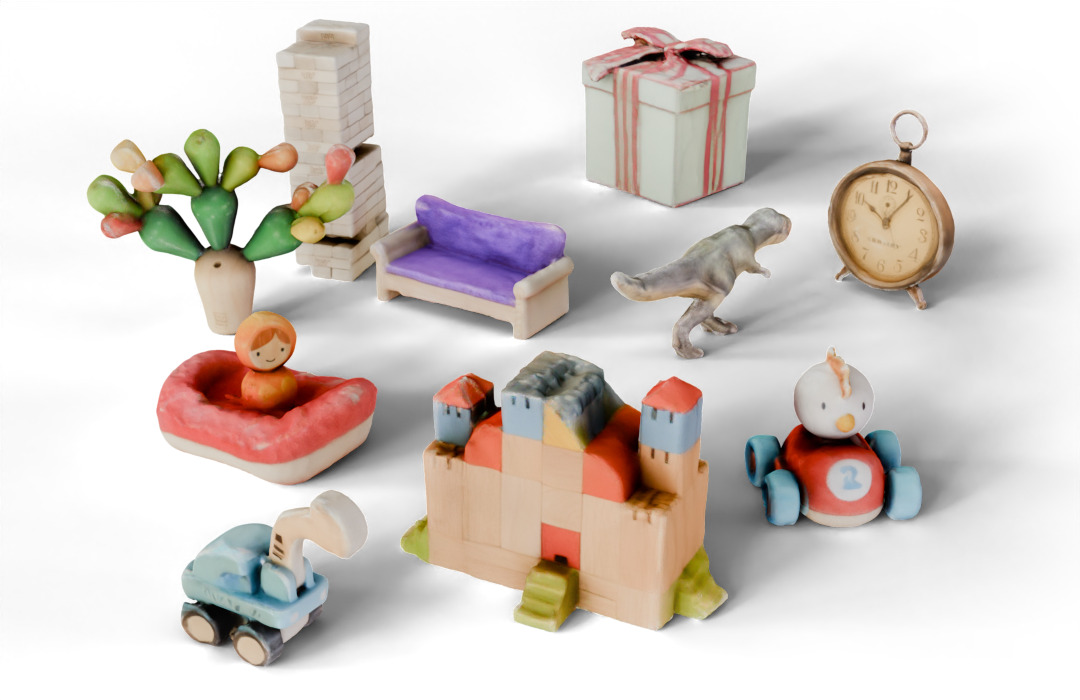}};
    \node at (opt_label -| examples) {\scriptsize Generated Meshes};
    \node at (opt_label -| film_strip) {\scriptsize Novel Multi-view Synthesis};

    \draw[-stealth] (input_img) -- (encoder);
    \draw[-stealth] (decoder.north) -| ($(film_strip.north east)+(-0.5,-0.1)$);
    \draw[-stealth, shorten <=-0.15cm] (film_strip.east) to [out=30,in=210] ($(nerf_viz)+(-0.075\textwidth,0.)$);
    \draw[-stealth] ($(nerf_viz)+(0.05\textwidth,0.0)$) --($(nerf_viz-|examples)+(-0.15\textwidth,0.cm)$);
\end{tikzpicture}
    \captionsetup{type=figure}
    \titlecaption{Stable Video 3D (SV3D)}{
        From a single image, SV3D generates consistent novel multi-view images. We then optimize a 3D representation with SV3D generated views resulting in high-quality 3D meshes. 
    }
    \label{fig:teaser}
    \end{center}

}]


\begin{NoHyper}
  \let\thefootnote\relax\footnotetext{* Core technical contribution.} 
\end{NoHyper}

\begin{NoHyper}
  \let\thefootnote\relax\footnotetext{Project Page: \url{https://sv3d.github.io/}} 
\end{NoHyper}

\begin{abstract}

We present Stable Video 3D (SV3D) --- a latent video diffusion model for high-resolution, image-to-multi-view generation of orbital videos around a 3D object. Recent work on 3D generation propose techniques to adapt 2D generative models for novel view synthesis (NVS) and 3D optimization. However, these methods have several disadvantages due to either limited views or inconsistent NVS, thereby affecting the performance of 3D object generation. In this work, we propose SV3D that adapts image-to-video diffusion model for novel multi-view synthesis and 3D generation, thereby leveraging the generalization and multi-view consistency of the video models, while further adding explicit camera control for NVS. We also propose improved 3D optimization techniques to use SV3D and its NVS outputs for image-to-3D generation. Extensive experimental results on multiple datasets with 2D and 3D metrics as well as user study demonstrate SV3D's state-of-the-art performance on NVS as well as 3D reconstruction compared to prior works.

\end{abstract}

\section{Introduction}
\label{sec:intro}

Single-image 3D object reconstruction is a long-standing problem in computer vision with a wide range of applications in game design, AR/VR, e-commerce, robotics, etc. It is a highly challenging and ill-posed problem as it requires lifting 2D pixels to 3D space while also reasoning about the unseen portions of the object in 3D.

Despite being a long-standing vision problem, it is only recently that practically useful results are being produced by leveraging advances in the generative AI. 
This is mainly made possible by the large-scale pretraining of generative models which enables sufficient generalization to various domains.
A typical strategy is to use image-based 2D generative models (\eg, Imagen~\cite{saharia2022photorealistic}, Stable Diffusion (SD)~\cite{rombach2022high})
to provide a 3D optimization loss function for unseen novel views of a given object~\cite{poole2022dreamfusion,metzer2023latentnerf,lin2023magic3d}.
In addition, several works repurpose these 2D generative models to perform novel view synthesis (NVS) from a single image~\cite{liu2023zero1to3,liu2023syncdreamer,watson2022novel,mercier2024hexagen3d}, and then use the synthesized novel views for 3D generation. 
Conceptually, these works mimic a typical photogrammetry-based 3D object capture pipeline, \ie, first photographing multi-view images of an object, followed by 3D optimization; except that the explicit multi-view capture is replaced by novel-view synthesis using a generative model, either via text prompt or camera pose control.

A key issue in these generation-based reconstruction methods is the lack of multi-view consistency in the underlying generative model resulting in inconsistent novel views. 
Some works~\cite{liu2023syncdreamer,chan2023genvs,hong2024lrm} try to address the multi-view consistency by jointly reasoning a 3D representation during NVS, but this comes at the cost of high computational and data requirements, often still resulting in unsatisfactory results with inconsistent geometric and texture details.
In this work, we tackle these issues by adapting a high-resolution, image-conditioned video diffusion model for NVS followed by 3D generation.

\inlinesection{Novel Multi-view Synthesis}. We adapt a latent video diffusion model (Stable Video Diffusion - SVD~\cite{blattmann2023stable}) to generate multiple novel views of a given object with \textit{explicit camera pose conditioning}.
SVD demonstrates excellent \textit{multi-view consistency} for video generation, and we repurpose it for NVS.
In addition, SVD also has good \textit{generalization} capabilities as it is trained on large-scale image and video data that are more readily available than large-scale 3D data. 
In short, we adapt the video diffusion model for NVS from a single image with three useful properties for 3D object generation: \textit{pose-controllable, multi-view consistent, and generalizable}.
We call our resulting NVS network `SV3D'.
To our knowledge, this is the first work that adapts a video diffusion model for explicit pose-controlled view synthesis. Some contemporary works such as~\cite{blattmann2023stable,melas2024im3d} demonstrate the use of video models for view synthesis, but they typically only generate orbital videos without any explicit camera control.

\inlinesection{3D Generation}.
We then utilize our SV3D model for 3D object generation by optimizing a NeRF and DMTet mesh in a coarse-to-fine manner. Benefiting from the multi-view consistency in SV3D, we are able to produce high-quality 3D meshes directly from the SV3D novel view images. We also design a masked score distillation sampling (SDS)~\cite{poole2022dreamfusion} loss to further enhance 3D quality in the regions that are not visible in the SV3D-predicted novel views. In addition, we propose to jointly optimize a disentangled illumination model along with 3D shape and texture, effectively reducing the issue of baked-in lighting.

We perform extensive comparisons of both our NVS and 3D generation results with respective state-of-the-art methods, demonstrating considerably better outputs with SV3D.
For NVS, SV3D shows high-level of multi-view consistency and generalization to real-world images while being pose controllable. Our resulting 3D meshes are able to capture intricate geometric and texture details.

\section{Background}
\label{sec:background}

\subsection{Novel View Synthesis}

We organize the related works along three crucial aspects of novel view synthesis (NVS): generalization, controllability, and multi-view (3D) consistency.

\inlinesection{Generalization.}
Diffusion models~\cite{ho2020ddpm,song2020sgm} have recently shown to be powerful generative models that can generate a wide variety of images~\cite{blattmann2023align,ruiz2022dreambooth,rombach2022high} and videos~\cite{blattmann2023stable,voleti2022mcvd,EMU} by iteratively denoising a noise sample.
Among these models, the publicly available Stable Diffusion (SD)~\cite{rombach2022high} and Stable Video Diffusion (SVD)~\cite{blattmann2023stable} demonstrate strong generalization ability by being trained on extremely large datasets like LAION~\cite{schuhmann2022laion} and LVD~\cite{blattmann2023stable}.
Hence, they are commonly used as foundation models for various generation tasks, \eg novel view synthesis.

\inlinesection{Controllability.}
Ideally, a controllable NVS model allows us to generate an image corresponding to any arbitrary viewpoint.
For this, Zero123~\cite{liu2023zero1to3} repurposes an image diffusion model to a novel view synthesizer, conditioned on a single-view image and the pose difference between the input and target views.
Follow-up works such as Zero123XL~\cite{deitke2023objaversexl} and Stable Zero123~\cite{stablezero123} advance the quality of diffusion-based NVS, as well as the trained NeRFs using SDS loss. However, these methods only generate one novel view at a time, and thus are not designed to be multi-view consistent inherently. Recent works such as EscherNet~\cite{kong2024eschernet} and Free3D~\cite{zheng2023free3d} are capable of generating with intelligent camera position embedding design encouraging better multi-view consistency. However, they only make use of image-based diffusion models, and generate images at 256$\times$256 resolution. We finetune a video diffusion model to generate novel views at 576$\times$576 resolution.

\inlinesection{Multi-view Consistency.}
Multi-view consistency is the most critical requirement for high-quality NVS and 3D generation.
To address the inconsistency issue in prior works, MVDream~\cite{shi2023mvdream}, SyncDreamer~\cite{liu2023syncdreamer}, HexGen3D~\cite{mercier2024hexagen3d}, and Zero123++~\cite{shi2023zero123++} propose to generate multiple (specific) views of an object simultaneously. However, they are not controllable: they only generate specific views given a conditional image, not arbitrary viewpoints. 
Moreover, they were finetuned from image-based diffusion models, \ie multi-view consistency was imposed on image-based diffusion by adding interaction among the multiple generated views through cross-attention layers. 
Hence, their output quality is limited to the generalization capability of the image-based model they finetuned from, as well as the 3D dataset they were finetuned on. Efficient-3DiM finetunes the SD model with a stronger vision transformer DINO v2~\cite{oquab2023dinov2}.
Consistent-1-to-3~\cite{ye2023consistent1to3} and SPAD leverage epipolar geometry to generate multiview consistent images.
One-2-3-45~\cite{liu2023one2345} and One-2-3-45++~\cite{liu2023one2345pp} train additional 3D network using the outputs of 2D generator. 
MVDream~\cite{shi2023mvdream}, Consistent123~\cite{weng2023consistent123}, and Wonder3D~\cite{long2023wonder3d} also train multi-view diffusion models, yet still require post-processing for video rendering.
SyncDreamer~\cite{liu2023syncdreamer} and ConsistNet~\cite{yang2023consistnet} employ 3D representation into the latent diffusion model.

\inlinesection{Exploiting Video Diffusion Models.}
To achieve better generalization and multi-view consistency, some contemporary works exploit the temporal priors in video diffusion models for NVS.
For instance, Vivid-1-to-3~\cite{kwak2023vivid} combines a view-conditioned diffusion model and video diffusion model to generate consistent views.
SVD-MV~\cite{blattmann2023stable} and IM-3D~\cite{melas2024im3d} finetune a video diffusion model for NVS. 
However, they generate $\leq$ 360$^\circ$ views at the same elevation only. 
Unlike SV3D, none of them are capable of rendering any arbitrary view of the 3D object.

We argue that the existing NVS and 3D generation methods do not fully leverage the superior generalization capability, controllability, and consistency in video diffusion models.
In this paper, we fill this important gap and train SV3D, a state-of-the-art novel multi-view synthesis video diffusion model at 576$\times$576 resolution, and leverage it for 3D generation.

\subsection{3D Generation}

Recent advances in 3D representations and diffusion-based generative models have significantly improved the quality of image-to-3D generation. Here, we briefly summarize the related works in these two categories.

\inlinesection{3D Representation.}
3D generation has seen great progress since the advent of Neural Radiance Fields (NeRFs)~\cite{mildenhall2020nerf} and its subsequent variants~\cite{barron2021mip}, which implicitly represents a 3D scene as a volumetric function, typically parameterized by a neural network. 
Notably, Instant-NGP~\cite{mueller2022instant} introduces a hash grid feature encoding that can be used as a NeRF backbone for fast inference and ability to recover complex geometry.
On the other hand, several recent works improve from an explicit representation such as DMTet~\cite{shen2021dmtet}, which is capable of generating high-resolution 3D shapes due to its hybrid SDF-Mesh representation and high memory efficiency.
Similar to~\cite{lin2023magic3d,qian2023magic123}, we adopt coarse-to-fine training for 3D generation, by first learning a rough object with Instant-NGP NeRF and then refining it using the DMTet representation.

\inlinesection{Diffusion-Based 3D Generation.}
Several recent works~\cite{jun2023shape,nichol2022pointe} train a 3D diffusion model to to learn these flexible 3D representations, which, however, lack generalization ability due to the scarcity of 3D data. 
To learn 3D generation without ground truth 3D data, image/multi-view diffusion models have been used as guidance for 3D generation.
DreamFusion~\cite{poole2022dreamfusion} and its follow-up works~\cite{metzer2023latentnerf,lin2023magic3d} leverage a trained image diffusion model as a `scoring' function and calculate the SDS loss for text-to-3D generation.
However, they are prone to artifacts like Janus problem~\cite{poole2022dreamfusion,metzer2023latentnerf} and over-saturated texture.
Inspired by Zero123~\cite{liu2023zero1to3}, several recent works~\cite{stablezero123,shi2023zero123++,liu2023syncdreamer,kong2024eschernet,zheng2023free3d,melas2024im3d,liu2023one2345,liu2023one2345pp,qian2023magic123} finetune image or video diffusion models to generate novel view images as a stronger guidance for 3D generation.
Our method shares the same spirit as this line of work, but produces denser, controllable, and more consistent multi-view images, thus resulting in better 3D generation quality.

\begin{figure*}[htb!]
\begin{minipage}[htb!]{.7\textwidth}
    \centering
    \input{figures/overview}
    \titlecaption{SV3D Architecture}{We add the sinusoidal embedding of the camera orbit elevation and azimuth angles $(e, a)$ to that of the noise step $t$, and feed the sum to the convolutional blocks in the UNet. We feed the single input image's CLIP embedding to the attention blocks, and concatenate its latent embedding to the noisy state $\rvz_t$.}
    \label{fig:sv3d_arch}
\end{minipage}\hfill
\begin{minipage}[htb!]{.27\textwidth}
    \centering
    \includegraphics[width=\linewidth,trim={20, 120, 10, 70},clip]{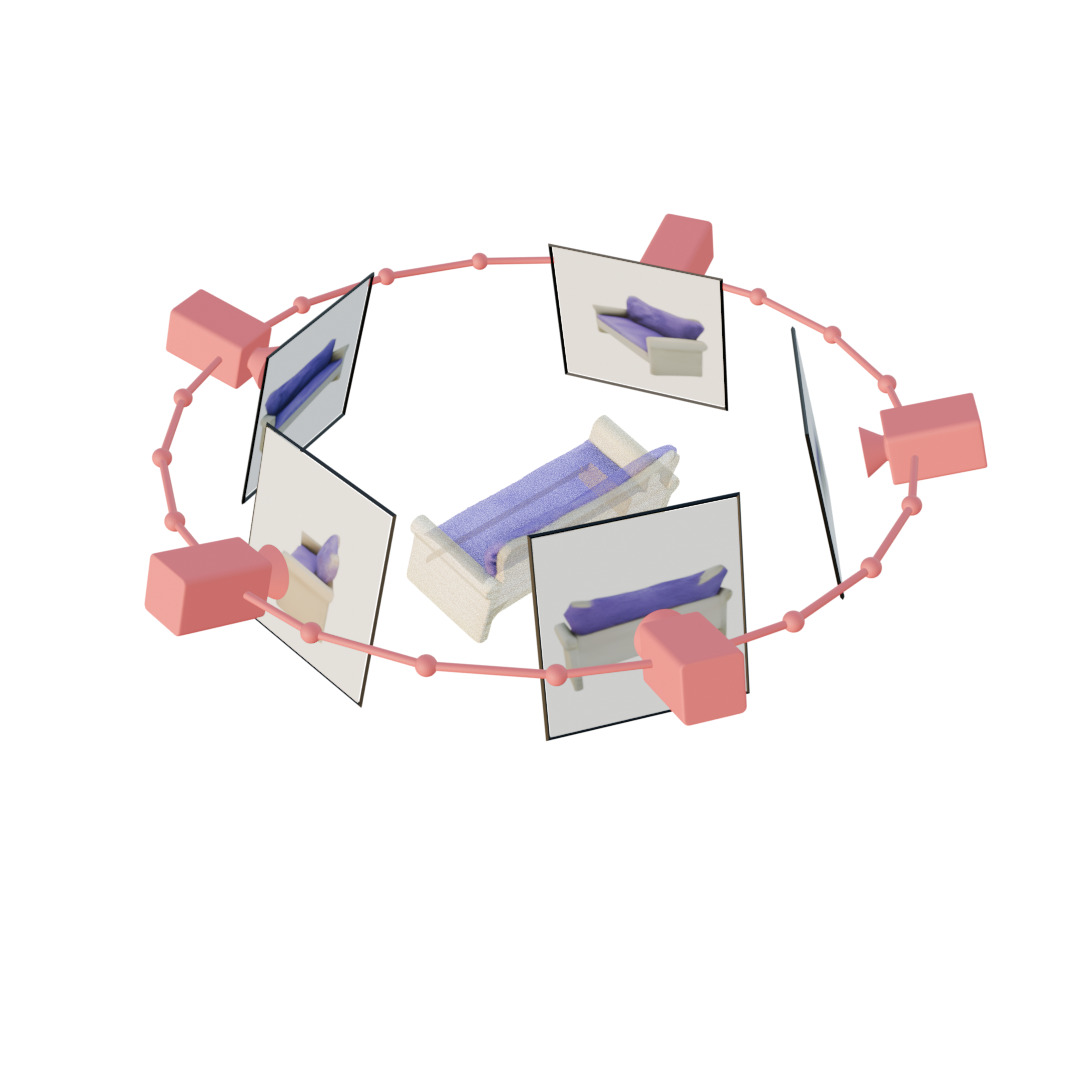} \\
    \scriptsize Static\\[0.15cm]
    \includegraphics[width=\linewidth,trim={20, 75, 10, 70},clip]{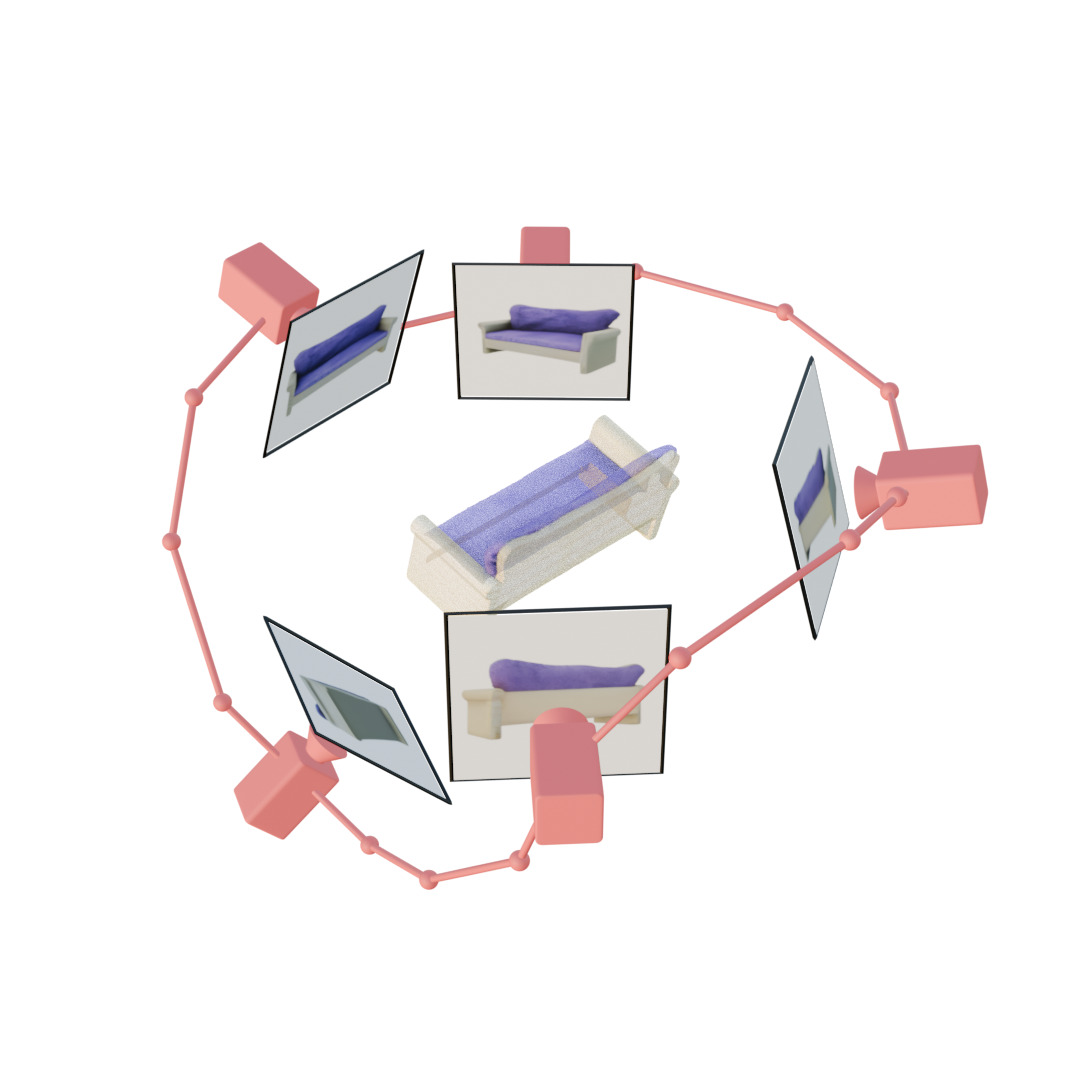} \\
    \scriptsize Dynamic
    \titlecaption{Static vs. Dynamic Orbits}{We use two types of orbits for training the SV3D models.}
    \label{fig:orbit_overview}
\end{minipage}
\end{figure*}

\section{SV3D: Novel Multi-view Synthesis}
\label{sec:nvs}

Our main idea is to repurpose temporal consistency in a video diffusion model for spatial 3D consistency of an object. Specifically, we finetune SVD~\cite{blattmann2023stable} to generate an orbital video around a 3D object, conditioned on a single-view image. This orbital video need not be at the same elevation, or at regularly spaced azimuth angles. SVD is well-suited for this task since it is trained to generate smooth and consistent videos on large-scale datasets of real and high-quality videos. The exposure to superior data quantity and quality makes it more \textit{generalizable} and \textit{multi-view consistent}, and the flexibility of the SVD architecture makes it amenable to be finetuned for \textit{camera controllability}. 

Some prior works attempt to leverage such properties by finetuning image diffusion models, training video diffusion models from scratch, or finetuning video diffusion models to generate pre-defined views at the same elevation (static orbit) around an object~\cite{blattmann2023stable,melas2024im3d}. However, we argue that these methods do not fully exploit the potential of video diffusion models. 
To the best of our knowledge, SV3D is the first video diffusion-based framework for \textit{controllable} multi-view synthesis at 576$\times$576 resolution (and subsequently for 3D generation).

\inlinesection{Problem Setting.} 
Formally, given an image ${\bf I} \in \mathbb{R}^{3 \times H \times W}$ of an object, our goal is to generate an orbital video ${\bf J} \in \mathbb{R}^{K \times 3 \times H \times W}$ around the object consisting of $K=21$ multi-view images along a camera pose trajectory $\bm{\pi} \in \mathbb{R}^{K \times 2} = \{(e_i, a_i)\}_{i=1}^K$ as a sequence of K tuples of elevation $e$ and azimuth $a$ angles. We assume that the camera always looks at the center of an object (origin of the world coordinates), and so any viewpoint can be specified by only two parameters: elevation and azimuth. We generate this orbital video by iteratively denoising samples from a learned conditional distribution $p({\bf J} | {\bf I}, \bm{\pi})$, parameterized by a video diffusion model.

\inlinesection{SV3D Architecture.}  As shown in \cref{fig:sv3d_arch}, the architecture of SV3D builds on that of SVD consisting of a UNet with multiple layers, each layer containing sequences of one residual block with Conv3D layers, and two transformer blocks (spatial and temporal) with attention layers.
(i) We remove the vector conditionings of `fps id' and `motion bucket id' since they are irrelevant for SV3D.
(ii) The conditioning image is concatenated to the noisy latent state input $\rvz_t$ at noise timestep $t$ to the UNet, after being embedded into latent space by the VAE encoder of SVD.
(iii) The CLIP-embedding~\cite{radford2021clip} matrix of the conditioning image is provided to the cross-attention layers of each transformer block as its key and value, the query being the feature at that layer. 
(iv) The camera trajectory is fed into the residual blocks along with the diffusion noise timestep. The camera pose angles $e_i$ and $a_i$ and the noise timestep $t$ are first embedded into sinusoidal position embeddings. Then, the camera pose embeddings are concatenated together, linearly transformed, and added to the noise timestep embedding. This is fed to every residual block, where they are added to the block's output feature (after being linearly transformed again to match the feature size).

\inlinesection{Static v.s. Dynamic Orbits.} 
As shown in \cref{fig:orbit_overview}, we design \textit{static} and \textit{dynamic} orbits to study the effects of camera pose conditioning.
In a \textbf{static} orbit, the camera circles around an object at regularly-spaced azimuths at the same elevation angle as that in the conditioning image. The disadvantage with the static orbit is that we might not get any information about the top or bottom of the object depending on the conditioning elevation angle. In a \textbf{dynamic} orbit, the azimuths can be irregularly spaced, and the elevation can vary per view. To build a dynamic orbit, we sample a static orbit, add small random noise to the azimuth angles, and add a random weighted combination of sinusoids with different frequencies to its elevation. This provides temporal smoothness, and ensures that the camera trajectory loops around to end at the same azimuth and elevation as those of the conditioning image.

Thus, with this strategy, we are able to tackle all three aspects of \textit{generalization}, \textit{controllability}, and \textit{consistency} in novel multi-view synthesis by leveraging video diffusion models, providing the camera trajectory as additional conditioning, and repurposing temporal consistency for spatial 3D consistency of the object, respectively.

\begin{figure*}[htb!]
\begin{minipage}[b]{.6\textwidth}
    \centering
    \input{figures/triangle_linear_cfg_scaling}
    \titlecaption{Linear vs. Triangle CFG Scaling}{Notice the increased oversharping in the penultimate frame in the linear scaling vs. our proposed triangle scaling.
    }
    \label{fig:cfg_triangle_linear_scale}
\end{minipage}\hfill
\begin{minipage}[b]{.38\textwidth}
  \begin{center}
    \pgfplotstableread[col sep = comma]{%
0.063,0.060,0.090,0.057,0.077,0.035
0.097,0.097,0.097,0.092,0.112,0.054
0.128,0.126,0.110,0.120,0.128,0.071
0.146,0.151,0.131,0.141,0.151,0.085
0.161,0.173,0.139,0.157,0.178,0.094
0.175,0.178,0.161,0.170,0.198,0.098
0.176,0.176,0.170,0.172,0.196,0.100
0.166,0.170,0.154,0.168,0.191,0.103
0.157,0.169,0.159,0.157,0.177,0.101
0.142,0.160,0.155,0.145,0.159,0.096
0.138,0.155,0.160,0.142,0.148,0.093
0.147,0.156,0.165,0.152,0.156,0.098
0.162,0.168,0.170,0.162,0.174,0.104
0.167,0.174,0.154,0.167,0.183,0.106
0.168,0.169,0.157,0.164,0.184,0.103
0.167,0.165,0.151,0.158,0.187,0.097
0.157,0.160,0.141,0.151,0.182,0.091
0.134,0.134,0.127,0.129,0.152,0.081
0.101,0.100,0.109,0.099,0.109,0.064
0.064,0.062,0.085,0.059,0.060,0.042
0.007,0.007,0.007,0.018,0.018,0.000
}\lpipsdata
\resizebox{\linewidth}{!}{
\begin{tikzpicture}[font=\scriptsize]
  \begin{axis}[
    legend cell align={left},
    legend style={at={(0.5,0.05)},anchor=south, nodes={scale=0.9, transform shape}},
    legend columns = 2, 
    xmin = 0, xmax = 20,
    ymin = 0, ymax = 0.21,
    height = 7.5cm,
    width = 7cm,
    yticklabel style={
        /pgf/number format/fixed,
        /pgf/number format/precision=3
    },
    scaled y ticks=false,
    xlabel = Frame,
    ylabel = LPIPS $\downarrow$,
    xtick pos=left,
    ytick pos=left,
    tick align=outside,
    xlabel near ticks,
    ylabel near ticks,
    axis on top,
    every x tick/.style={color=black, thin},
    every y tick/.style={color=black, thin},
    minor x tick num=1,
    cycle list/Dark2,
    every axis plot/.append style={ultra thick}
    ]
    \addplot table[x expr=\coordindex,y index=0]{\lpipsdata};
    \addplot table[x expr=\coordindex,y index=1]{\lpipsdata};
    \addplot table[x expr=\coordindex,y index=2]{\lpipsdata};
    \addplot table[x expr=\coordindex,y index=3]{\lpipsdata};
    \addplot table[x expr=\coordindex,y index=4]{\lpipsdata};
    \addplot table[x expr=\coordindex,y index=5]{\lpipsdata};
    \legend{Zero123,Zero123XL,Stable Zero123,EscherNet,Free3D,\textbf{SV3D$^p$}};
  \end{axis}
\end{tikzpicture}}
  \end{center}
  \vspace{-6mm}
  \titlecaption{LPIPS vs. Frame Number}{We find that SV3D has the best reconstruction metric per frame.}
    \label{fig:lpips}
\end{minipage}
\end{figure*}

\inlinesection{Triangular CFG Scaling.} 
SVD uses a linearly increasing scale for classifier-free guidance (CFG) from 1 to 4 across the generated frames. However, this scaling causes the last few frames in our generated orbits to be over-sharpened, as shown in \cref{fig:cfg_triangle_linear_scale} Frame 20.
Since we generate videos looping back to the front-view image, we propose to use a triangle wave CFG scaling during inference: linearly increase CFG from 1 at the front view to 2.5 at the back view, then linearly decrease it back to 1 at the front view.
\cref{fig:cfg_triangle_linear_scale} also demonstrates that our triangle CFG scaling produces more details in the back view (Frame 12).

\inlinesection{Models.}
We train three image-to-3D-video models finetuned from SVD. First, we train a pose-\textit{u}nconditioned model, $\oursu$, which generates a video of static orbit around an object while only conditioned on a single-view image. Note that unlike SVD-MV~\cite{blattmann2023stable}, we do not provide the elevation angle to the pose-unconditioned model, as we find that the model is able to infer it from the conditioning image. Our second model, the pose-\textit{c}onditioned $\oursc$ is conditioned on the input image as well as a sequence of camera elevation and azimuth angles in an orbit, trained on dynamic orbits. Following SVD's~\cite{blattmann2023stable} intuition to \textit{p}rogressively increase the task difficulty during training, we train our third model, $\oursp$, by first finetuning SVD to produce static orbits unconditionally, then further finetuning on dynamic orbits with camera pose condition.

\inlinesection{Training Details.}
We train SV3D on the Objaverse dataset~\cite{deitke2023objaverse}, which contains synthetic 3D objects covering a wide diversity.
For each object, we render 21 frames around it on random color background at 576$\times$576 resolution, field-of-view of 33.8 degrees.
We choose to finetune the SVD-xt model to output 21 frames.
All three models ($\oursu$, $\oursc$, $\oursp$) are trained for 105k iterations in total ($\oursp$ is trained unconditionally for 55k iterations and conditionally for 50k iterations), with an effective batch size of 64 on 4 nodes of 8 80GB A100 GPUs for around 6 days.
For more training details, please see the appendix.

\subsection{Experiments and Results}
\label{subsec:nvs_eval}

\inlinesection{Datasets.} We evaluate the SV3D synthesized multi-view images on static and dynamic orbits on the unseen GSO~\cite{downs2022google} and OmniObject3D~\cite{wu2023omniobject3d} datasets. Since many GSO objects are the same items with slightly different colors, we filter 300 objects from GSO to reduce redundancy and maintain diversity. For each object, we render ground truth static and dynamic orbit videos and pick the last frame of each video as the conditioning image. We also conduct a user study on novel view videos from a dataset of 22 real-world images. See the appendix for more details on the datasets.

\inlinesection{Metrics.} We use the SV3D models to generate static and dynamic orbit videos corresponding to the ground truth camera trajectories in the evaluation datasets. We compare each generated frame with the corresponding ground truth frames, in terms of Learned Perceptual Similarity (LPIPS~\cite{zhang2018lpips}), Peak Signal-to-Noise Ratio (PSNR), Structural SIMilarity (SSIM), Mean Squared-Error (MSE), and CLIP-score (CLIP-S). This range of metrics covers both pixel-level as well as semantic aspects. Note that testing on dynamic orbits evaluates the controllability of SV3D models, and all the metrics evaluate multi-view consistency.

\inlinesection{Baselines.} We compare SV3D with several recent NVS methods capable of generating arbitrary views, including Zero123~\cite{liu2023zero1to3}, Zero123-XL~\cite{deitke2023objaversexl}, SyncDreamer \cite{liu2023syncdreamer}, Stable Zero123~\cite{stablezero123}, Free3D~\cite{zheng2023free3d}, EscherNet~\cite{kong2024eschernet}.

\inlinesection{Results.} 
As shown in \cref{tab:nvs-GSO-static,tab:nvs-GSO-dynamic,tab:nvs-OO3D-static,tab:nvs-OO3D-dynamic}, our SV3D models achieve state-of-the-art performance on novel multi-view synthesis. \cref{tab:nvs-GSO-static,tab:nvs-OO3D-static} show results on static orbits, and include all our three models. We see that even our pose-unconditioned model $\oursu$ performs better than all prior methods. \cref{tab:nvs-GSO-dynamic,tab:nvs-OO3D-dynamic} show results on dynamic orbits, and include our pose-conditioned models $\oursc$ and $\oursp$.

\NVScomp

\inlinesection{Ablative Analyses.} Interestingly, from \cref{tab:nvs-GSO-static,tab:nvs-OO3D-static}, we find that both $\oursc$ and $\oursp$ outperform $\oursu$ on generations of static orbits, even though $\oursu$ is trained specifically on static orbits. 
We also observe that $\oursp$ achieves better metrics than $\oursc$ on both static and dynamic orbits, making it the best performing SV3D model overall.
This shows that progressive finetuning from easier (static) to harder (dynamic) tasks is indeed a favorable way to finetune a video diffusion model.

\begin{table}[htb!]
\caption{Evaluation of novel multi-view synthesis on GSO static orbits}
\label{tab:nvs-GSO-static}
\centering
\resizebox{\linewidth}{!}{
\begin{tabular}{ l c c c c c }
\toprule 
 Model & LPIPS$\downarrow$ & PSNR$\uparrow$ & SSIM$\uparrow$ & CLIP-S$\uparrow$ & MSE$\downarrow$ \\
 \midrule
SyncDreamer~\cite{liu2023syncdreamer}  & 0.17 & 15.78 & 0.76 & 0.87 & 0.03          \\
Zero123~\cite{liu2023zero1to3}         & 0.13 & 17.29 & 0.79 & 0.85 & 0.04          \\
Zero123XL~\cite{deitke2023objaversexl} & 0.14 & 17.11 & 0.78 & 0.85 & 0.04          \\
Stable Zero123~\cite{stablezero123}    & 0.13 & 18.34 & 0.78 & 0.85 & 0.05          \\
Free3D~\cite{zheng2023free3d}          & 0.15          & 16.18          & 0.79    & 0.84          & 0.04          \\
EscherNet~\cite{kong2024eschernet}     & 0.13          & 16.73          & 0.79    & 0.85          & 0.03          \\
\midrule
$\oursu$ & 0.09          & 21.14          & 0.87 & \textbf{0.89} & \textbf{0.02}          \\
$\oursc$            & 0.09 & 20.56 & 0.87 & 0.88 & \textbf{0.02} \\
\textbf{$\oursp$}  & \textbf{0.08} & \textbf{21.26} & \textbf{0.88} & \textbf{0.89} & \textbf{0.02} \\
\bottomrule
\end{tabular}
}
\end{table}
\begin{table}[htb!]
\caption{Evaluation of novel multi-view synthesis on GSO dynamic orbits}
\label{tab:nvs-GSO-dynamic}
\centering
\resizebox{\linewidth}{!}{
\begin{tabular}{ l c c c c c }
\toprule 
 Model & LPIPS$\downarrow$ & PSNR$\uparrow$ & SSIM$\uparrow$ & CLIP-S$\uparrow$ & MSE$\downarrow$ \\
 \midrule
Zero123~\cite{liu2023zero1to3}            & 0.14          & 16.99          & 0.79          & 0.84          & 0.04          \\
Zero123XL~\cite{deitke2023objaversexl}          & 0.14          & 16.73          & 0.78          & 0.84          & 0.04          \\
Stable Zero123~\cite{stablezero123}     & 0.13          & 18.04          & 0.78          & 0.85          & 0.05          \\
Free3D~\cite{zheng2023free3d}          &  0.18         &    14.93       & 0.77          & 0.83     &   0.05        \\
EscherNet~\cite{kong2024eschernet}          &  0.13         &    16.47       & 0.79          & 0.84     &   0.03        \\
\midrule
$\oursc$  & 0.10          & 19.99          & 0.86          & \textbf{0.87}          & \textbf{0.02} \\
\textbf{$\oursp$} & \textbf{0.09} & \textbf{20.38} & \textbf{0.87} & \textbf{0.87} & \textbf{0.02} \\
\bottomrule
\end{tabular}
}
\end{table}
\begin{table}[htb!]
\caption{Evaluation of novel multi-view synthesis on OmniObject3D static orbits
}
\label{tab:nvs-OO3D-static}
\centering
\resizebox{\linewidth}{!}{
\begin{tabular}{ l c c c c c }
\toprule 
 Model & LPIPS$\downarrow$ & PSNR$\uparrow$ & SSIM$\uparrow$ & CLIP-S$\uparrow$ & MSE$\downarrow$ \\
 \midrule
Zero123~\cite{liu2023zero1to3}             & 0.17          & 15.50          & 0.76          & 0.83          & 0.05          \\
Zero123XL~\cite{deitke2023objaversexl}           & 0.18          & 15.36          & 0.75          & 0.83          & 0.06          \\
Stable Zero123~\cite{stablezero123}      & 0.15          & 16.86          & 0.77          & 0.84          & 0.06          \\
Free3D~\cite{zheng2023free3d}              & 0.16	          & 15.29          & 	0.78	          & 0.83          & 0.05          \\
EscherNet~\cite{kong2024eschernet}           & 0.17          & 14.63          & 0.74          & 0.83          & 0.05          \\
\midrule
$\oursu$ & \textbf{0.10} & 19.68          & \textbf{0.86} & \textbf{0.86} & \textbf{0.02} \\
$\oursc$            & \textbf{0.10} & 19.50          & 0.85          & 0.85          & \textbf{0.02} \\
\textbf{$\oursp$}  & \textbf{0.10} & \textbf{19.91} & \textbf{0.86} & \textbf{0.86} & \textbf{0.02} \\
\bottomrule
\end{tabular}
}
\end{table}
\begin{table}[htb!]
\caption{Evaluation of novel multi-view synthesis on OmniObject3D dynamic orbits}
\label{tab:nvs-OO3D-dynamic}
\centering
\resizebox{\linewidth}{!}{
\begin{tabular}{ l c c c c c }
\toprule 
 Model & LPIPS$\downarrow$ & PSNR$\uparrow$ & SSIM$\uparrow$ & CLIP-S$\uparrow$ & MSE$\downarrow$ \\
 \midrule
Zero123~\cite{liu2023zero1to3}             & 0.16      & 15.78     & 0.77      & 0.82                & 0.05                \\
Zero123XL~\cite{deitke2023objaversexl}           & 0.17      & 15.49     & 0.76      & 0.83                & 0.05                \\
Stable Zero123~\cite{stablezero123}      & 0.14      & 16.74     & 0.77      & 0.83                & 0.05                \\
Free3D~\cite{zheng2023free3d}              & 0.19	& 14.28 & 0.76	 & 0.82 & 0.06 \\
EscherNet~\cite{kong2024eschernet}           & 0.16  & 15.05 & 0.76 & 0.83 & 0.05 \\
\midrule
$\oursc$           & \textbf{0.10}  & 19.21 & \textbf{0.85}  & 0.84  & \textbf{0.02}                \\
\textbf{$\oursp$}      & \textbf{0.10}  & \textbf{19.28} & \textbf{0.85}  & \textbf{0.85}  & \textbf{0.02} \\
\bottomrule
\end{tabular}
}
\end{table}

\inlinesection{Visual Comparisons} in \cref{fig:nvs_comp} further demonstrate that SV3D-generated images are more detailed, faithful to the conditioning image, and multi-view consistent compared to the prior works.

\inlinesection{Quality Per Frame.} We also observe from \cref{fig:lpips} that SV3D produces better quality at every frame. We plot the average LPIPS value for each generated frame, across generated GSO static orbit videos. The quality is generally worse around the back view, and better at the beginning and the end (\ie near the conditioning image), as expected.

\inlinesection{User Study on Real-World Images.} We conducted a user survey to study human preference between static orbital videos generated by SV3D and by other methods. We asked 30 users to pick one between our SV3D-generated static video and other method-generated video as the best orbital video for the corresponding image, using 22 real-world images. We noted that users preferred SV3D-generated videos over Zero123XL, Stable Zero123, EscherNet, and Free3D, 96\%, 99\%, 96\%, and 98\% of the time, respectively.

\section{3D Generation from a Single Image Using SV3D}

\begin{figure*}
    \centering
    \resizebox{.9\textwidth}{!}{
    \input{figures/recon_teaser}}
    \titlecaption{3D Optimization Pipeline}{We perform a two-stage optimization. In the short first stage, we extract the general shape, texture and illumination from the SV3D generated multi-view images. In the second stage, we extract a mesh with marching cubes and use DMTet to further optimize the shape, texture and illumination. We not only use the SV3D-generated images but a soft-masked SDS loss for unseen areas. Dashed red lines represent backpropagation into the differentiable rendering pipeline.}
    \label{fig:reconstruction_overview}
\end{figure*}

We then generate 3D meshes of objects from a single image by leveraging SV3D. One way is to use the generated static/dynamic orbital samples from SV3D as direct reconstruction targets. Another way is to use SV3D as diffusion guidance with Score Distillation Sampling (SDS) loss~\cite{poole2022dreamfusion}.

Since SV3D produces more consistent multi-views compared to existing works, we already observe higher-quality 3D reconstructions by only using SV3D outputs for reconstruction when compared to existing works.
However, we observe that this naive approach often leads to artifacts like baked-in illumination, rough surfaces, and noisy texture, especially for the unseen regions in the reference orbit.
Thus, we further propose several techniques to address these issues.

\inlinesection{Coarse-to-Fine Training.}
We adopt a two-stage, coarse-to-fine training scheme to generate a 3D mesh from input images, similar to~\cite{lin2023magic3d,qian2023magic123}.
\cref{fig:reconstruction_overview} illustrate an overview of our 3D optimization pipeline.
In the coarse stage, we train an Instant-NGP~\cite{mueller2022instant} NeRF representation to reconstruct the SV3D-generated images (\ie without SDS loss) at a lower resolution.
In the fine stage, we extract a mesh from the trained NeRF using marching cubes~\cite{cline1988admissibile}, and adopt a DMTet~\cite{shen2021dmtet} representation to finetune the 3D mesh using SDS-based diffusion guidance from SV3D at full-resolution.
Finally, we use xatlas~\cite{xatlas} to perform the UV unwrapping and export the output object mesh.

\illumination
\inlinesection{Disentangled Illumination Model.} %
Similar to other recent 3D object generation methods~\cite{poole2022dreamfusion,lin2023magic3d,metzer2023latentnerf}, our output target is a mesh with a diffuse texture.
Typically, such SDS-based optimization techniques use random illuminations at every iteration.
However, our SV3D-generated videos are under consistent illumination, \ie, the lighting remains static while the camera circles around an object.
Hence, to disentangle lighting effects and obtain a cleaner texture, we propose to fit a simple illumination model of 24 Spherical Gaussians (SG) inspired by prior decomposition methods~\cite{boss2021nerd, zhang2021physg}.
We model white light and hence only use a scalar amplitude for the SGs. We only consider Lambertian shading, where the cosine shading term is approximated with another SG. We learn the parameters of the illumination SGs using a reconstruction loss between the rendered images and SV3D-generated images.

\orbitsphotosds

Inspired by~\cite{poole2022dreamfusion,hasselgren2022nvdiffrecmc} we reduce baked-in illumination with a loss term that replicates the HSV-value component of the input image $\bf I$ with the rendered illumination $L$: $\mathcal{L}_\text{illum}=\left|V({\bf I}) - L\right|^2$, with $V(\bm{c})=\max(c_r,c_g,c_b)$.
Given these changes, our disentangled illumination model is able to express lighting variation properly and can severely reduce baked-in illumination. \cref{fig:diffuse_vs_sgs} shows sample reconstructions with and without our illumination modelling. From the results, it is clear that we are able to disentangle the illumination effects from the base color (e.g., dark side of the school bus).

\subsection{3D Optimization Strategies and Losses}
\inlinesection{Reconstruction via Photometric Losses.}
Intuitively, we can treat the SV3D-generated images as multi-view pseudo-ground truth, and apply 2D reconstruction losses to train the 3D models.
In this case, we apply photometric losses on the rendered images from NeRF or DMTet, including the pixel-level MSE loss, mask loss, and a perceptual LPIPS~\cite{zhang2018lpips} loss.
These photometric losses also optimize our illumination model through the differential rendering pipeline.

\inlinesection{Training Orbits.}
For 3D generation, we use SV3D to generate multi-view images following a camera orbit $\bm{\pi}_\text{ref}$, referred to as the \textit{reference orbit} (also see \cref{fig:reconstruction_overview} for the overview).
\cref{fig:orbits} shows sample reconstruction with using static and dynamic orbital outputs from SV3D. 
Using a dynamic orbit for training is crucial to high-quality 3D outputs when viewed from various angles, since some top/bottom views are missing in the static orbit (fixed elevation).
Hence, for $\oursc$ and $\oursp$, we render images on a dynamic orbit whose elevation follows a sine function to ensure that top and bottom views are covered.

\inlinesection{SV3D-Based SDS Loss.} In addition to the reconstruction losses, we can also use SV3D via score-distillation sampling (SDS)~\cite{poole2022dreamfusion,yao2024artic3d}. \cref{fig:photo_sds} shows sample reconstructions with and without using SDS losses. 
As shown in \cref{fig:photo_sds} (left), although training with a dynamic orbit improves overall visibility,
we observe that sometimes the output texture is still noisy, perhaps due to partial visibility, self-occlusions, or inconsistent texture/shape between images.
Hence, we handle those unseen areas using SDS loss~\cite{poole2022dreamfusion} with SV3D as a diffusion guidance.

We sample a random camera orbit $\bm\pi_\text{rand}$, and use our 3D NeRF/DMTet parameterized by $\theta$ to render the views ${\hat{\bf J}}$ of the object along $\bm\pi_\text{rand}$. Then, noise $\epsilon$ at level $t$ is added to the latent embedding $\rvz_t$ of ${\hat{\bf J}}$, and
the following SDS loss (taken expectation over $t$ and $\epsilon$) is backpropagated through the differentiable rendering pipeline: $\mathcal{L}_\text{sds} = w(t) \big( \epsilon_\phi({\bf z}_t; {\bf I}, \bm\pi_\text{rand}, t) - \epsilon \big) \frac{\partial \hat{\bf{J}}}{\partial \theta},$
where $w$ is $t$-dependent weight,
$\epsilon$ and $\epsilon_\phi$ are the added and predicted noise, 
and $\phi$ and $\theta$ are the parameters of SV3D and NeRF/DMTet, respectively. See~\cref{fig:reconstruction_overview} for an illustration of these loss functions in the overall pipeline.

\inlinesection{Masked SDS Loss.}
As shown in \cref{fig:photo_sds} (middle), in our experiments we found that adding the SDS loss naively can cause unstable training and unfaithful texture to the input images such as oversaturation or blurry artifacts.
Therefore, we design a soft masking mechanism to only apply SDS loss on the unseen/occluded areas, allowing it to inpaint the missing details while preserving the texture of clearly-visible surfaces in the training orbit (as seen in \cref{fig:photo_sds} (right)).
Also, we only apply the masked SDS loss in the final stage of DMTet optimization, which greatly increased the convergence speed.

We apply SDS loss on only those pixels in the random orbit views that are not likely visible in the reference orbit views. For this, we first render the object from the random camera orbit $\bm\pi_{\text{rand}}$. For each random camera view, we obtain the visible surface points $\bm{p} \in \mathbb{R}^3$ and their corresponding surface normals $\bm{n}$. 
Then, for each reference camera $i$, we calculate the view directions $\bm{v}_i$ of the surface $\bm{p}$ towards its position $\bm{\bar{\pi}}_\text{ref}^i \in \mathbb{R}^3$ (calculated from $\bm\pi_\text{ref}^i \in \mathbb{R}^2$) as $\bm{v}_i=\frac{\bm{\bar{\pi}}_\text{ref}^i - \bm{p}}{|| \bm{\bar{\pi}}_\text{ref}^i - \bm{p} ||}$.
We infer the visibility of this surface from the reference camera based on the dot product between $\bm{v}_i$ and $\bm{n}$ \ie $\bm{v}_i \cdot \bm{n}$. Since higher values roughly indicate more visibility of the surface from the reference camera, we chose that reference camera $c$ that has maximum likelihood of visibility: $c = \max_i \left( \bm{v}_i \cdot \bm{n} \right)$.
As we only want to apply SDS loss to unseen or grazing angle areas from $c$, we use the smoothstep function $f_s$ to smoothly clip to $c$'s visibility range $\bm{v}_c \cdot \bm{n}$. In this way, we create a pseudo visibility mask $M =1-f_\text{s} \left(\bm{v}_c \cdot \bm{n} , 0, 0.5\right)$, where $f_\text{s}(x;f_0,f_1)=\hat{x}^2 (3 - 2x)$, with $\hat{x}=\frac{x-f_0}{f_1-f_0}$. Thus, $M$ is calculated for each random camera render, and the combined visibility mask $\rmM$ is applied to SDS loss: $\mathcal{L}_\text{mask-sds}= \rmM \mathcal{L}_\text{sds}$.

\inlinesection{Geometric Priors.} 
Since our rendering-based optimization operates at the image level, we adopt several geometric priors to regularize the output shapes.
We add a smooth depth loss from RegNeRF~\cite{Niemeyer2021Regnerf} and a bilateral normal smoothness loss~\cite{boss2022-samurai} to encourage smooth 3D surfaces where the projected image gradients are low.
Moreover, we obtain normal estimates from Omnidata~\cite{Eftekhar2021omnidata} and calculate a mono normal loss similar to MonoSDF~\cite{Yu2022MonoSDF}, which can effectively reduce noisy surfaces in the output mesh.
Further details about the training losses and optimization process are available in the appendix.

\baselinethreed

\realthreed

\subsection{Experiments and Results}

Due to the strong regularization, we only require 600 steps in the coarse stage and 1000 in the fine stage.
Overall, the entire mesh extraction process takes $\approx$8 minutes without SDS loss, and $\approx$20 minutes with SDS loss.
The coarse stage only takes $\approx$2 minutes and provides a full 3D representation of the object.

\inlinesection{Evaluation.} We evaluate our 3D generation framework on 50 randomly sampled objects from the GSO dataset as described in \cref{subsec:nvs_eval}.
We compute image-based reconstruction metrics (LPIPS, PSNR, SSIM, MSE, and CLIP-S) between the ground truth (GT) GSO images, and rendered images from the trained 3D meshes on the same static/dynamic orbits.
In addition, we compute 3D reconstruction metrics of Chamfer distance (CD) and 3D IoU between the GT and predicted meshes.
We compare our SV3D-guided 3D generations with several prior methods including Point-E~\cite{nichol2022pointe}, Shap-E~\cite{jun2023shape}, One-2-3-45++~\cite{liu2023one2345pp}, DreamGaussian~\cite{tang2023dreamgaussian}, SyncDreamer~\cite{liu2023syncdreamer}, EscherNet~\cite{kong2024eschernet}, Free3D \cite{zheng2023free3d}, and Stable Zero123~\cite{stablezero123}.

\inlinesection{Visual Comparison.}
In \cref{fig:3d_baselines}, we show visual comparison of our results with those from prior methods.
Qualitatively, Point-E~\cite{nichol2022pointe} and Shap-E~\cite{jun2023shape} often produce incomplete 3D shapes.
DreamGaussian~\cite{tang2023dreamgaussian}, SyncDreamer~\cite{liu2023syncdreamer}, EscherNet~\cite{kong2024eschernet}, and Free3D~\cite{zheng2023free3d} outputs tend to contain rough surfaces.
One-2-3-45++~\cite{liu2023one2345pp} and Stable Zero123~\cite{stablezero123} are able to reconstruct meshes with smooth surface, but lack geometric details.
Our mesh outputs are detailed, faithful to input image, and consistent in 3D (see appendix for more examples).
We also show 3D mesh renders from real-world images in \cref{fig:3d_real}.

\inlinesection{Quantitative Comparison.} 
\cref{tab:threeD,tab:threeD2} show the 2D and 3D metric comparisons respectively. All our 3D models achieve better 2D and 3D metrics compared to the prior and concurrent methods, showing the high-fidelity texture and geometry of our output meshes.
We render all 3D meshes on the same dynamic orbits and compare them against the GT renders. 
Our best model, $\oursp$, performs comparably to using GT renders for reconstruction in terms of the 3D metrics, which further demonstrates the 3D consistency of our generated images.

\inlinesection{Effects of Photometric and (Masked) SDS Loss.}
As shown in \cref{tab:threeD,tab:threeD2}, the 3D outputs using both photometric and Masked SDS losses (`\textbf{$\oursp$}') achieves the best results, while training without SDS (`$\oursp$ no SDS') leads to marginally lower performance.
This demonstrates that the images generated by SV3D are high-quality reconstruction targets, and are often sufficient for 3D generation without the cumbersome SDS-based optimization.
Nevertheless, adding SDS generally improves quality, as also shown in \cref{fig:photo_sds}.

\inlinesection{Effects of SV3D Model and Training Orbit.}
As shown in \cref{tab:threeD,tab:threeD2}, $\oursp$ achieves the best performance among the three SV3D models, indicating that its synthesized novel views are most faithful to the input image and consistent in 3D.
On the other hand, $\oursu$ shares the same disadvantage as several prior works in that it can only generate views at the same elevation, which is insufficient to build a legible 3D object, as shown in \cref{fig:orbits}.
This also leads to the worse performance of `$\oursp$ with static orbit' in \cref{tab:threeD,tab:threeD2}.
Overall, $\oursp$ with dynamic orbit and masked SDS loss performs favorably against all other configurations since it can leverage more diverse views of the object.

\inlinesection{Limitations.} Our SV3D model is by design only capable of handling 2 degrees of freedom: elevation and azimuth; which is usually sufficient for 3D generation from a single image. One may want to tackle more degrees of freedom in cameras for a generalized NVS system, which forms an interesting future work.
We also notice that SV3D exhibits some view inconsistency for mirror-like reflective surfaces, which provide a challenge to 3D reconstruction. Lastly, such reflective surfaces are not representable by our Lambertian reflection-based shading model. Conditioning SV3D on the full camera matrix, and extending the shading model are interesting directions for future research.

\begin{table}[t!]
\centering
\caption{2D comparison of our 3D outputs against prior methods on the GSO dataset. Our best performing method uses \textbf{$\oursp$} with dynamic (sine elevation) orbit and SDS guidance. Note that all our models achieve better 2D metrics than prior works.}
\label{tab:threeD}
\resizebox{\linewidth}{!}{
\begin{tabular}{ l c c c c c}
\toprule
 Model & LPIPS$\downarrow$ & PSNR$\uparrow$ & SSIM$\uparrow$ & MSE$\downarrow$ & CLIP-S$\uparrow$ \\
 \midrule
 GT renders                          & 0.078 & 19.508 & 0.879 & 0.014 & 0.897 \\
 \midrule
 EscherNet~\cite{kong2024eschernet}  & 0.178 & 14.438 & 0.804 & 0.041 & 0.835 \\ 
 Free3D~\cite{zheng2023free3d}       & 0.197 & 14.202 & 0.799 & 0.043 & 0.809 \\ 
 Stable Zero123~\cite{stablezero123} & 0.166 & 14.635 & 0.813 & 0.040 & 0.805 \\
 \midrule
 $\oursu$                            & 0.133 & 15.957 & 0.834 & 0.031 & 0.871 \\
 $\oursc$                            & 0.132 & 16.373 & 0.834 & 0.027 & 0.870 \\
 $\oursp$ static orbit               & 0.125 & 16.821 & 0.848 & 0.025 & 0.864 \\
 $\oursp$ no SDS                     & 0.124 & 16.864 & 0.841 & 0.024 & 0.875 \\
 \textbf{$\oursp$}                   & \bf0.119 & \bf17.405 & \bf0.849 & \bf0.021 & \bf0.877 \\
 \bottomrule
\end{tabular}}
\end{table}
\begin{table}[t!]
\centering
\caption{Comparison of 3D metrics. Our models perform favorably against prior works.}
\label{tab:threeD2}
\resizebox{0.6\linewidth}{!}{
\begin{tabular}{ l c c }
\toprule 
 Model & CD$\downarrow$ & 3D IoU$\uparrow$ \\
 \midrule
 GT renders                                 & 0.021 & 0.689 \\
 \midrule
 Point-E~\cite{nichol2022pointe}            & 0.074 & 0.162 \\
 Shap-E~\cite{jun2023shape}                 & 0.071 & 0.267 \\
 DreamGaussian~\cite{tang2023dreamgaussian} & 0.055 & 0.411 \\
 One-2-3-45++~\cite{liu2023one2345pp}             & 0.054 & 0.406 \\
 SyncDreamer~\cite{liu2023syncdreamer}      & 0.053 & 0.451 \\
 EscherNet~\cite{kong2024eschernet}         & 0.042 & 0.466 \\
 Free3D~\cite{zheng2023free3d}              & 0.047 & 0.426 \\
 Stable Zero123~\cite{stablezero123}        & 0.039 & 0.550 \\
 \midrule
 $\oursu$                                   & 0.027 & 0.589 \\
 $\oursc$                                   & 0.027 & 0.584 \\
 $\oursp$ static orbit                      & 0.028 & 0.610 \\
 $\oursp$ no SDS                            & \bf0.024 & 0.611 \\
 \textbf{$\oursp$}                          & \bf0.024 & \bf0.614 \\
 \bottomrule
\end{tabular}}
\end{table}

\section{Conclusion}

We present SV3D, a latent video diffusion model for novel multi-view synthesis and 3D generation. In addition to leveraging the generalizability and view-consistent prior in SVD, SV3D enables controllability via camera pose conditioning, and generates orbital videos of an object at high resolution on arbitrary camera orbits. We further propose several techniques for improved 3D generation from SV3D, including triangle CFG scaling, disentangled illumination, and masked SDS loss. We conduct extensive experiments to show that SV3D is controllable, multi-view consistent, and generalizable to the real-world, achieving state-of-the-art performance on novel multi-view synthesis and 3D generation. We believe SV3D provides a solid foundation model for further research on 3D object generation. We plan to publicly release SV3D models.

\vspace{1mm}
\noindent \textbf{Acknowledgments}.
 We thank Emad Mostaque, Bryce Wilson, Anel Islamovic, Savannah Martin, Ana Guillen, Josephine Parquet, Adam Chen and Ella Irwin for their help in various aspects of the model development and release. We thank Kyle Lacey and Yan Marek for their help with demos. We also thank Eric Courtemanche for his help with visual results.

{
    \small
    \bibliographystyle{ieeenat_fullname}
    \bibliography{arxiv,non_arxiv,postings,old}
}

\newpage
\onecolumn

\tableofcontents
\appendix
\section*{Appendix}





\section{Broader Impact}
\label{sec:impact}
The advancement of generative models in different media forms is changing how we make and use content. These AI-powered models can create images, videos, 3D objects, and more, in ways we've never seen before. They offer huge potential for innovation in media production.
But along with this potential, there are also risks. Before we start using these models widely, it is crucial to make sure we understand the possible downsides and have plans in place to deal with them effectively.

In the case of 3D object generation, the input provided by the user plays a crucial role in constraining the model's output. By supplying a full front view of an object, users limit the model's creative freedom to the visible or unoccluded regions, thus minimizing the potential for generating problematic imagery. Additionally, factors such as predicted depth values and lighting further influence the fidelity and realism of generated content.

Moreover, ensuring the integrity and appropriateness of training data is critical in mitigating risks associated with generative models. Platforms like Sketchfab, which serve as repositories for 3D models used in training, enforce strict content policies to prevent the dissemination of “Unacceptable Content” and disallows it on their platform: https://help.sketchfab.com/hc/en-us/articles/214867883-What-is-Restricted-Content. By adhering to these guidelines and actively monitoring dataset quality, developers can reduce the likelihood of biased or inappropriate outputs.

Sketchfab also has a tag for “Restricted Content” which is deemed to be similar to the PG-13 content rating used in the US (i.e. inappropriate for children under 13).  We have confirmed that none of the objects that we use in training have this flag set to true. Thus we go the extra step of excluding even tagged PG-13 content from the training set.

There is a chance that certain content may not have been correctly labeled on Sketchfab. In cases where the uploader fails to tag an object appropriately, Sketchfab provides publicly accessible listings of objects and a mechanism for the community to report any content that may be deemed offensive.

In our regular utilization of Objaverse content, we haven't observed any significant amount of questionable material. Nevertheless, there are occasional instances of doll-like nudity stemming from basic 3D models, which could be crucial for accurately depicting humanoid anatomy. Additionally, the training dataset contains some presence of drugs, drug paraphernalia, as well as a certain level of blood content and weaponry, resembling what might be encountered in a video game context. Should the model be provided with imagery featuring these categories of content, it possesses the capability to generate corresponding 3D models to some extent.

It is to be noted that SV3D mainly focuses on generating hidden details in the user's input image. If the image is unclear or some parts are hidden, the model guesses what those parts might look like based on its training data. This means it might create details similar to what it has seen before. The training data generally follows the standards of 3D modeling and gaming. However, this could lead to criticisms about the models being too similar to existing trends. But the user's input image limits how creative the model can be and reduces the chance of biases showing up in its creations, especially if the image is clear and straightforward.

\section{Data Details}
\label{sec:data}
Similar to SVD-MV~\cite{blattmann2023stable}, we render views of a subset of the Objaverse dataset~\cite{deitke2023objaverse} consisting of 150K curated CC-licensed 3D objects from the original dataset. Each loaded object is scaled such that the largest world-space XYZ extent of its bounding box is 1. The object is then repositioned such that it this bounding box is centered around the origin.

For both the static and dynamic orbits, we use Blender's EEVEE renderer to render an 84-frame RGBA orbit at 576$\times$576 resolution. During training, any 21-frame orbit can be subsampled from this by picking any frame as the first frame, and then choosing every 4th frame after that.

We apply two background colors to each of these images: random RGB color, and white. This results in a doubling of the number of orbit samples for training. We then encode all of these images into latent space using SVD's VAE, and using CLIP. We then store the latent and CLIP embeddings for all of these images along with the corresponding elevation and azimuth values.

For lighting, we randomly select from a set of 20 curated HDRI envmaps. Each orbit begins with the camera positioned at azimuth 0. Our camera uses a field-of-view of 33.8 degrees. For each object, we adaptively position the camera to a distance sufficient to ensure that the rendered object content makes good and consistent use of the image extents without being clipped in any view.

For static orbits, the camera is positioned at a randomly sampled elevation between [-5, 30] degrees. The azimuth steps by a constant $\frac{360}{84}$ degree delta between each frame.
For dynamic orbits, the seqeunce of camera elevations for each orbit are obtained from a random weighted combination of sinusoids with different frequencies. For each sinusoid, the whole number period is sampled from [1, 5], the amplitude is sampled from [0.5, 10], and a random phase shift is applied.
The azimuth angles are sampled regularly, and then a small amount of noise is added to make them irregular.
The elevation values are smoothed using a simple convolution kernel and then clamped to a maximum elevation of 89 degrees.

\section{Training Details}
\label{sec:training}
Our approach involves utilizing the widely used EDM~\cite{karras2022elucidating} framework, incorporating a simplified diffusion loss for fine-tuning, as followed in SVD~\cite{blattmann2023stable}. We eliminate the conditions of `fps\_id', `motion\_bucket\_id', etc. since they are irrelevant for SV3D. Furthermore, we adjust the loss function to assign lower weights to frames closer to the front-view conditioning image, ensuring that challenging back views receive equal training focus as the easier front views. To optimize training efficiency and conserve GPU VRAM, we preprocess and store the precomputed latent and CLIP embeddings for all video frames in advance. During training, these tensors are directly loaded rather than being computed in real-time. We choose to finetune the SVD-xt model to output 21 frames instead of 25 frames. We found that with 21 frames we were able to fit a batch size of 2 on each GPU, instead of 1 with 25 frames at 576$\times$576 resolution. All SV3D models are trained for 105k iterations with an effective batch size of 64 on 4 nodes of 8 80GB A100 GPUs for around 6 days.

\section{Inference Details}
\label{sec:inference}
To generate an orbital video during inference, we use 50 steps of the deterministic DDIM sampler~\cite{song2020improved} with the triangular classifier-free guidance (CFG) scale described in the main paper. This takes $\approx$40 seconds for the SV3D model. 

\section{Additional Details on Illumination Model}
\label{sec:lighting}
We base our rendering model on Spherical Gaussians (SG)~\cite{boss2021nerd, zhang2021physg}.
A SG at query location $\bm{x} \in \mathbb{R}^3$ is defined as $G(\bm{x};\bm{\mu},c,a)=a e^{s (\bm{\mu} \cdot \bm{x} -1)}$, where $\bm{\mu} \in \mathbb{R}^3$ is the axis, $s \in \mathbb{R}$ the sharpness of the lobe, and $a \in \mathbb{R}$ the amplitude. Here, we point out that we only model white light and hence only use a scalar amplitude.
One particularly interesting property of SGs is that the inner product of two SGs is the integral of the product of two SGs. 
The operation is defined as~\cite{tsai-sg2006}:
\begin{align}
    G_1(\bm{x}) \cdot G_2(\bm{x}) &=\int_\Omega G_1(\bm{x}) G_2(\bm{x}) d\bm{x} = %
    \frac{1}{d_m} \left(2 \pi a_1 a_2 %
    e^{d_m-\lambda_m} %
    (1.0 - e^{-2 d_m}) \right) \nonumber \\
    \lambda_m &= \lambda_1 - \lambda_2 \\
    d_m &= || \lambda_1 \bm{\mu}_1 + \lambda_2 \bm{\mu}_2 || \nonumber .
\end{align}

In our illumination model we only consider Lambertian shading. Here, the cosine shading term influences the output the most. This term can be approximated with another SG $G_\text{cosine}=(\bm{x}; \bm{n},2.133,1.17)$, where $\bm{n}$ defines the surface normal at $\bm{x}$.
The lighting evaluation using SGs $G_i$ then becomes: $L= \sum^{24}_{i=1} \frac{1}{\pi} \max(G_i \cdot G_\text{cosine}, 0)$. 
As defined previously this results in the full integration of incoming light for each SG and as light is additive evaluating and summing all SGs results in the complete environment illumination.
This $L$ is also used in the $\mathcal{L}_\text{illum}$ loss described in the main paper. 
The rendered textured image is then defined as $\bm{\hat{I}}=\bm{c_d} \bm{L}$, where $\bm{c_d}$ is the diffuse albedo. We learn $\bm{\mu},c,a$ for each SG $G_i$ using reconstruction loss between these rendered images and SV3D-generated images.

\section{Losses and Optimization for 3D Generation}
\label{sec:losses}
In addition to the masked SDS loss $\mathcal{L}_\text{mask-sds}$ and illumination loss $\mathcal{L}_\text{illum}$ detailed in the manuscript, we use several other losses for 3D reconstruction. 
Our main reconstruction losses are the pixel-level mean squared error $\mathcal{L}_\text{mse} = \lVert \bm{I}-\bm{\hat{I}} \rVert^2$, LPIPS~\cite{zhang2018lpips} loss $\mathcal{L}_\text{lpips}$, and mask loss $\mathcal{L}_\text{mask} = \lVert \bm{S} - \bm{\hat{S}} \rVert^2$, where $\bm{S}$, $\bm{\hat{S}}$ are the predicted and ground-truth masks.
%
We further employ a normal loss using the estimated mono normal by Omnidata~\cite{Eftekhar2021omnidata}, which is defined as the cosine similarity between the rendered normal $\bm{n}$ and estimated pseudo ground truths $\bm{\bar{n}}$: $\mathcal{L}_\text{normal}=1-(\bm{n} \cdot \bm{\bar{n}})$.
To regularize the output geometry, 
we apply a smooth depth loss inspired by RegNeRF~\cite{Niemeyer2021Regnerf}:
$\mathcal{L}_\text{depth}(i,j)=\left(d(i,j)-d(i,j+1)\right)^2 + \left(d(i,j)-(d(i+1,j)\right)^2$, where $i,j$ indicate the pixel coordinate.
For surface normal we instead rely on a bilateral smoothness loss similar to~\cite{boss2022-samurai}. We found that this is crucial to getting high-frequency details and avoiding over-smoothed surfaces. For this loss we compute the image gradients of the input image $\nabla \bm{I}$ with a Sobel filter~\cite{kanopoulos1988sobel}. We then encourage the gradients of rendered normal $\nabla \bm{n}$ to be smooth if (and only if) the input image gradients $\nabla \bm{I}$ are smooth. The loss can be written as $\mathcal{L}_\text{bilateral}=e^{-3 \nabla \bm{I}} \sqrt{1 + || \nabla \bm{n} ||}$.
We also found that adding a spatial smoothness regularization on the albedo is beneficial: $\mathcal{L}_\text{albedo} = \left( \bm{c}_d(\bm{x}) - \bm{c}_d(\bm{x} + \bm{\epsilon})\right)^2$, where $\bm{c}_d$ denotes the albedo, $\bm{x}$ is a 3D surface point, and $\bm{\epsilon} \in \mathbb{R}^3$ is a normal distributed offset with a scale of $0.01$.
The overall objective is then defined as the weighted sum of these losses. All losses are applied in both coarse and fine stages, except that we only apply $\mathcal{L}_\text{mask-sds}$ in the last 200 iterations of the fine stage.
%
%
%
We use an Adam optimizer~\cite{kingma2014adam} with a learning rate of $0.01$ for both stages.

\section{Additional Ablative Analyses}
\label{sec:ablation}
We conduct additional ablative analyses of our 3D generation pipeline in this section.

\subsection{SV3D Models}
\label{subsec:models}
In \cref{tab:3d_supp}, we compare the quantitative results using different SV3D models and training losses. Both 2D and 3D evaluation shows that $\oursp$ is our best performing model, either for pure photometric reconstruction or SDS-based optimization.

\begin{table*}[!]
\centering
\small
\caption{{\bf Ablative results of different SV3D models and training losses.} We show that our $\oursp$ model with Photo+SDS losses achieves the best 2D and 3D metrics.}
\begin{tabular}{ l l c c c c c c c }
\label{tab:3d_supp}
 Model     & Training losses & LPIPS$\downarrow$ & PSNR$\uparrow$ & SSIM$\uparrow$ & MSE$\downarrow$ & CLIP-S$\uparrow$ & CD$\downarrow$ & 3D IoU$\uparrow$ \\
 \hline
 $\oursu$  & Photo     & 0.132 & 15.951 & 0.827 & 0.032 & 0.873 & 0.028 & 0.583 \\
 $\oursu$  & Photo+SDS & 0.133 & 15.957 & 0.834 & 0.031 & 0.871 & 0.027 & 0.589 \\
 $\oursc$  & Photo     & 0.135 & 15.826 & 0.832 & 0.033 & 0.871 & 0.029 & 0.579 \\
 $\oursc$  & Photo+SDS & 0.132 & 16.373 & 0.834 & 0.027 & 0.870 & 0.027 & 0.584 \\
 $\oursp$  & Photo     & 0.124 & 16.864 & 0.841 & 0.024 & 0.875 & \bf0.024 & 0.611 \\
 $\oursp$  & Photo+SDS & \bf0.119 & \bf17.405 & \bf0.849 & \bf0.021 & \bf0.877 & \bf0.024 & \bf0.614 \\
\end{tabular}
\end{table*}

\subsection{Static v.s. Dynamic Orbits}
\label{subsec:orbits}
We also compare the results using different camera orbits for 3D training in \cref{tab:3d_supp2}. The results show that using a dynamic orbit (sine-30) produces better 3D outputs compared to static orbit since it contains more information of the top and bottom views of the object. However, higher elevation (sine-50) tends to increase inconsistency between multi-view images, and thus resulting in worse 3D reconstruction. In our experiments, we find that setting the elevation within $\pm30$ degree generally leads to desirable 3D outputs.

\begin{table*}[!]
\centering
\small
\caption{{\bf Ablative results of different reference orbits for 3D generation.} We show that using a dynamic orbit (sine elevation) with moderate amplitude performs better than orbits with no or extreme elevation variations.}
\begin{tabular}{ c c c c c c c c }
\label{tab:3d_supp2}
 Training orbit & LPIPS$\downarrow$ & PSNR$\uparrow$ & SSIM$\uparrow$ & MSE$\downarrow$ & CLIP-S$\uparrow$ & CD$\downarrow$ & 3D IoU$\uparrow$ \\
 \hline
 Static    & 0.125 & 16.821 & 0.848 & 0.025 & 0.864 & 0.028 & 0.610 \\
 Sine-{30} & \bf0.119 & \bf17.405 & 0.849 & \bf0.021 & \bf0.877 & \bf0.024 & \bf0.614 \\
 Sine-{50} & 0.123 & 17.057 & \bf0.854 & 0.025 & 0.873 & 0.026 & 0.609 \\
\end{tabular}
\end{table*}

\subsection{Masked SDS Loss}
\label{subsec:sds}
Finally, we show the ablative results of our SDS loss in \cref{tab:3d_supp3}. We compare the results of 1) pure photometric losses, 2) with naive SDS loss (no masking), 3) with hard-masked SDS loss by thresholding the dot product of surface normal and camera viewing angle as visibility masks, and 4) with soft-masked SDS loss as described in the manuscript. Overall, adding SDS guidance from the SV3D model can improve the 2D metrics while maintaining similar 3D metrics. Our novel soft-masked SDS loss generally achieves the best results compared to other baselines.

\begin{table*}[!]
\centering
\small
\caption{{\bf Ablative analyses of Masked SDS loss.} Overall, our soft-masked SDS loss leads to higher-quality mesh outputs in terms of most 2D and 3D metrics.}
\begin{tabular}{ l c c c c c c c }
\label{tab:3d_supp3}
 Training losses & LPIPS$\downarrow$ & PSNR$\uparrow$ & SSIM$\uparrow$ & MSE$\downarrow$ & CLIP-S$\uparrow$ & CD$\downarrow$ & 3D IoU$\uparrow$ \\
 \hline
 Photo                   & 0.124 & 16.864 & 0.841 & 0.024 & 0.875 & \bf0.024 & 0.611 \\
 Photo+SDS (naive)       & 0.124 & 17.007 & \bf0.850 & 0.024 & 0.867 & 0.025 & \bf0.615 \\
 Photo+SDS (hard masked) & 0.124 & 17.335 & 0.845 & 0.022 & \bf0.877 & \bf0.024 & 0.610 \\
 Photo+SDS (soft masked) & \bf0.119 & \bf17.405 & 0.849 & \bf0.021 & \bf0.877 & \bf0.024 & 0.614 \\
\end{tabular}
\end{table*}

\section{Additional Visual Results}
\label{sec:results}
In this section, we show more results of novel view synthesis and 3D generation.

\subsection{Novel View Synthesis}
\label{subsec:nvs}
We show the additional NVS results on OmniObject3D~\cite{wu2023omniobject3d} and real-world images in \cref{fig:nvs_omni_supp} and \cref{fig:nvs_real_supp}, respectively. The generated novel multi-view images by SV3D are more detailed and consistent compared to prior state-of-the-arts.

\omniobjectnvssupp

\realnvssupp

\subsection{3D Generation}
\label{subsec:threeDgen}
We show the additional 3D generation results on OmniObject3D~\cite{wu2023omniobject3d} and real-world images in \cref{fig:3d_omni_supp} and \cref{fig:3d_real_supp}, respectively. Thanks to the consistent multi-view images by SV3D and the novel Masked SDS loss, our 3D generations are detailed, high-fidelity, and generalizable to a wide range of images. Since Free3D~\cite{zheng2023free3d} does not include a 3D generation method, we run our 3D pipeline on its generated multi-view images for fair 3D comparion.

\omniobjectthreedsupp

\realthreedsupp

\end{document}